\documentclass{article}
\PassOptionsToPackage{numbers, compress}{natbib}

\usepackage[preprint]{neurips_2026}

\usepackage[utf8]{inputenc}
\usepackage[T1]{fontenc}
\usepackage{hyperref}
\usepackage{url}
\usepackage{booktabs}
\usepackage{amsfonts}
\usepackage{amsmath}
\usepackage{amssymb}
\usepackage{nicefrac}
\usepackage{microtype}
\usepackage[dvipsnames]{xcolor}
\usepackage{xspace}
\usepackage{graphicx}
\usepackage{pgfplots}
\pgfplotsset{compat=1.18}
\usepackage{placeins}  
\usepackage{cleveref}
\usepackage{enumitem}
\usepackage{listings}

\lstdefinestyle{walkgraph}{
  basicstyle=\scriptsize\ttfamily,
  keywordstyle=\bfseries,
  commentstyle=\itshape\color{gray},
  stringstyle=\color{gray},
  showstringspaces=false,
  columns=fullflexible,
  frame=single,
  framesep=2pt,
  xleftmargin=2pt,
  breaklines=true,
  language=Python,
  morekeywords={Sequential,Parallel,Loop,DynamicLoop,Node,Edge,StreamingGraphEdge,FixedChunkPolicy,SlidingWindowChunkPolicy,LeftContextChunkPolicy,EMIT_TO_CLIENT}
}

\newif\ifcomments
\commentsfalse
\ifcomments
    \newcommand{\TODO}[1]{\textcolor{red}{\textbf{[TODO:} #1\textbf{]}}}
    
    \newcommand{\baris}[1]{\textcolor{purple}{\textbf{[BK:} #1\textbf{]}}}
    \providecommand{\swang}[1]{{\protect\color{cyan}{\bf [swang: #1]}}}
    \providecommand{\keisuke}[1]{{\protect\color{orange}{\bf [keisuke: #1]}}}
    \providecommand{\naomi}[1]{{\protect\color{blue}{\bf [naomi: #1]}}}
    
    \providecommand{\atindra}[1]{{\protect\color{magenta}{\bf [atindra: #1]}}}
    \providecommand{\owh}[1]{{\protect\color{ForestGreen}{ [\textbf{owh}: #1]}}}
\else
    \newcommand{\baris}[1]{}
    \providecommand{\swang}[1]{}
    \providecommand{\keisuke}[1]{}
    \providecommand{\TODO}[1]{}
    \providecommand{\atindra}[1]{}
    \providecommand{\naomi}[1]{}
    \providecommand{\owh}[1]{}
\fi

\newcommand{\mminf}{\textsc{M*}\xspace}    
\newcommand{\sys}{\mminf}

\title{\sys{}: A Modular, Extensible, Serving System\\
       for Multimodal Models}

\author{%
  \bfseries Atindra Jha$^{1,*}$ \quad Naomi Sagan$^{1,*}$ \quad Keisuke Kamahori$^{2,\dagger}$ \quad Irmak Sivgin$^{1,\dagger}$ \\
  \bfseries Rohan Sanda$^{1}$ \quad Steven Gao$^{2}$ \quad Mark Horowitz$^{1}$ \quad Luke Zettlemoyer$^{2}$ \\
  \bfseries Olivia Hsu$^{1,3}$ \quad Jure Leskovec$^{1,\ddagger}$ \quad Baris Kasikci$^{2,\ddagger}$ \quad Stephanie Wang$^{2,\ddagger}$ \\[5pt]
  {\normalfont\small $^{1}$Stanford University \quad $^{2}$University of Washington \quad $^{3}$Carnegie Mellon University} \\[3pt]
  {\normalfont\small $^{*}$Co-first authors \quad $^{\dagger}$Second authors \quad $^{\ddagger}$Equal advising} \\[3pt]
  {\normalfont\small Correspondence to \texttt{atindra@cs.stanford.edu}}
}

\begin{document}
\maketitle

\begin{abstract}
We are entering a new era of {\em composite model architectures} that integrate diverse components such as vision encoders, language backbones, diffusion and flow heads, audio codecs, action generators, and world-model predictors.
Such architectures underpin a broad class of multimodal models, including unified multimodal models, omni models, speech-language models, vision-language-action policies, and world models. However, existing model serving frameworks were built on narrow assumptions about model structure, making them ill-suited to accommodate this new architectural diversity.
Here we present \sys{}, a universal serving system for efficient serving of composite AI models.
\sys{} represents models as dataflow graphs, processing requests spanning diverse modalities and tasks as traversals over these graphs. The core insight is a modular abstraction that supports arbitrary composition of model components, flexible placement onto a physical cluster, and model-agnostic optimizations within a distributed runtime.  We call this abstraction the \textbf{Walk Graph} and show how it can concisely capture composite models from a broad range of families.
We instantiate \sys{} on representative models and find that it achieves, on average, 20\% lower end-to-end latency than vLLM-Omni for text-to-image workloads on BAGEL, while delivering up to 2.9$\times$ lower real-time factor and 2.7$\times$ higher throughput for text-to-speech workloads on Qwen3-Omni.
\sys{} also outperforms the V-JEPA 2-AC rollout baseline for robotic planning by up to 12.5$\times$.
Thus, our work paves the road towards more efficient serving of complex models with minimal developer effort.
\end{abstract}

\section{Introduction}
\label{sec:intro} 
AI is entering a new era of \emph{composite model architectures}: multimodal models built from structurally distinct components, including vision encoders, transformer backbones, diffusion and flow heads, audio codecs, and action generators.
Unlike earlier models with relatively fixed execution structures, these components are composed and executed in patterns that vary across inputs and tasks.
This new generation includes unified multimodal models (UMMs), omni models, speech language models (SpeechLMs), vision-language-action models (VLAs), and world models~\citep{deng2025bagel,xu2025qwen3omni,orpheus2025,black2025pi05,assran2025vjepa2}.
Tasks span image and video understanding and generation, real-time speech interaction, robot interaction, and world prediction. Despite their differences in modality and task, these architectures share a key property: \emph{inference no longer reduces to a single autoregressive forward loop}.

The diverse execution structures of multimodal models create requirements that current LLM serving stacks~\cite{kwon2023vllm,zheng2024sglang} do not cleanly capture.
In text-only LLMs, all data takes essentially the same path through a simple autoregressive (AR) loop.
Meanwhile, in modern multimodal LLMs, different modalities and tasks may trigger different execution paths through the same heterogeneous model.
For example, in UMMs such as BAGEL~\citep{deng2025bagel}, image generation vs. image understanding tasks pass data through different components within the same heterogeneous model.
Other models may contain long non-AR loops, such as diffusion transformers (DiTs)~\citep{peebles2023dit} or rectified flow~\citep{liu2023rectified, esser2024sd3} for image generation and variable-horizon world-model rollouts~\citep{assran2025vjepa2}.
They may also contain internal parallelism, such as the condition branches in classifier-free guidance (CFG)~\citep{ho2022cfg} or the pipelined Thinker--Talker architecture in Qwen3-Omni~\citep{xu2025qwen3omni}.
These patterns are not isolated exceptions layered onto otherwise token-centric models but are becoming the standard structure of multimodal models.

Despite rapid progress, modern LLM serving systems~\cite{kwon2023vllm,zheng2024sglang} still face a fundamental abstraction mismatch when extended from AR-focused text generation to composite multimodal inference. 
These frameworks have been successfully adapted to multimodal models that attach a non-text encoder to a language model backbone, as in vision-language models (VLMs) or speech recognition models~\citep{bai2025qwen3vl,radford2023robust}.
However, they are insufficient in capturing the complex patterns described above.

Recent work has attempted to address this gap with an intermediate abstraction of a fixed chain~\cite{yin2026vllmomni} or DAG~\cite{sglangomni} of ``stages'', where each stage captures one or many model components.
However, a gap still remains for modern multimodal models, resulting in suboptimal performance.
For example, the stage abstraction cannot capture more complex patterns such as non-AR loops across stages, parallelism internal to a stage, and different tasks or modalities taking different paths through the same model (see \S\ref{sec:background}).
Thus, the underlying system misses key performance opportunities, such as parallelism across components or request-specific execution of components.
Furthermore, physical placement can only be controlled at the granularity of a stage, which misses efficiency opportunities such as independent scaling of individual components.

This work addresses the problem of building an efficient multimodal serving system that enables day-zero support for the next generation of composite models. Our key insight is that, despite the diversity of model architectures, \emph{every multimodal model is a dataflow graph of heterogeneous components}, and every user request executes as a \emph{walk} of components within the graph.
We design a flexible intermediate abstraction based on this idea to decouple the model architecture from the system runtime, enabling: (1) the capture of diverse model architectures, and (2) the support for flexible placement of different components to maximize hardware utilization, while (3) achieving same or better performance as custom-built serving engines.

Thus, we present \sys{}, a universal multimodal serving system.
After the model author declares their model architecture as a computation graph and a set of graph walks, the deployer instantiates the model by declaring a mapping of model components to physical GPU ranks.
Then, the runtime is responsible for all physical execution, such as component disaggregation, request scheduling, request batching, tensor transport, and tensor streaming.
The runtime also integrates well-established optimizations, including paged attention~\citep{kwon2023vllm}, CUDA-graph capture, and continuous batching~\citep{yu2022orca}, as well as state-of-the-art modality-specific optimizations.

By decoupling the model architecture from the system runtime, \sys{} enables broad support for composite multimodal models spanning text, image, video, audio, and robot actions.
\sys{} maintains state-of-the-art per-component performance and further improves efficiency because \emph{the system execution mirrors the model's component graph}.
We demonstrate these capabilities by instantiating \sys{} on five representative composite models: BAGEL, Qwen3-Omni, $\pi_{0.5}$, V-JEPA~2, and Orpheus.
On BAGEL~\citep{deng2025bagel}, \sys{} delivers, on average, \textasciitilde20\% lower p50 end-to-end latency than vLLM-Omni~\citep{yin2026vllmomni} on text-to-image generation and up to $2.64\times$ lower on image editing (with CFG parallelism enabled), while improving image-understanding (I2T) throughput by up to 32.7\% for 64--256-token outputs---serving all three workloads with a single configuration, whereas each vLLM-Omni configuration is tuned for only a subset.
On Qwen3-Omni text-to-speech, \sys{} consistently delivers lower RTF and higher throughput than both vLLM-Omni and SGLang-Omni~\cite{sglangomni} across batch sizes, reaching up to $2.9\times$ lower RTF and, at batch size~16, $2.7\times$ and $4.0\times$ higher throughput than vLLM-Omni and SGLang-Omni, respectively.
On Orpheus-TTS~\citep{orpheus2025}, \sys{} outperforms VoxServe~\cite{kamahori2026voxserve}, a speech-optimized serving system, with $13.6\%$ lower p50 RTF and $39\%$ higher throughput at batch size~16.
Finally, \sys{} accelerates V-JEPA~2~\citep{assran2025vjepa2} robotic-planning rollouts by up to $12.5\times$ over the native implementation.  

\section{Background and Motivation}
\label{sec:background}

\subsection{The Era of Composite Multimodal Models}
\label{sec:compound-models}

Unlike text-only LLMs, recent multimodal models exhibit substantial architectural diversity: heterogeneous components for consuming and producing data across modalities, connected in complex structures. We describe five representative families to exemplify the structural patterns that we target.

\begin{figure}[t]
  \centering
  \includegraphics[width=0.8\linewidth,page=4,trim={0cm 4cm 4cm 0cm},clip]{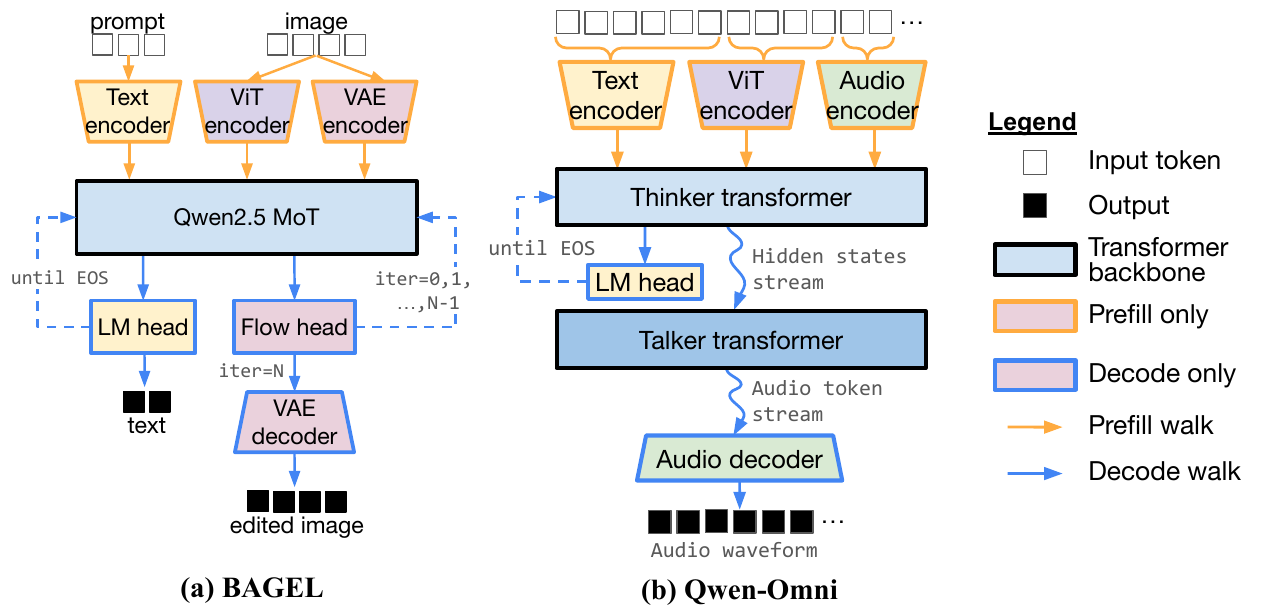}
  \caption{Example model architectures of (a) a UMM (BAGEL~\citep{deng2025bagel}) and (b) Omni model (Qwen3-Omni~\citep{xu2025qwen3omni}). Composite models are structurally diverse, yet representable as computation graphs.
}
  \label{fig:model-taxonomy}
\end{figure}

\begin{enumerate}[leftmargin=*, itemsep=2pt, topsep=2pt, parsep=0pt]
    \item \textbf{Unified multimodal models (UMMs)} are designed to handle multiple types of vision tasks with a shared Transformer backbone and encoders/decoders for text and visual latents~\citep{deng2025bagel,chen2025januspro,xie2025showo2}. For example, BAGEL~\citep{deng2025bagel} handles vision understanding, image generation, and image editing in a single model with a shared Mixture-of-Transformers (MoT~\citep{liang2024mot}) component. Each task uses a different combination of encoders/decoders and transformer weights~(\Cref{fig:model-taxonomy}a).

    \item \textbf{Speech language models (SpeechLMs)} pair an autoregressive Transformer backbone with neural audio codec models to serve text-to-speech or speech-to-speech applications~\citep{orpheus2025,du2024cosyvoice2,huang2025stepaudio}.
    The codec model needs to be invoked at different intervals to generate audio waveforms depending on the architecture. For streaming applications, output audio must be produced in real time~\citep{kamahori2026voxserve}.
    \item \textbf{Omni models} aim to support any-to-any modalities, including real-time speech. Qwen2.5-Omni and Qwen3-Omni models~\citep{xu2025qwen25omni,xu2025qwen3omni} are notable examples that combine two Transformers in a Thinker--Talker topology. The Thinker produces text and high-level hidden states from inputs in any modality, while the Talker converts those states into speech codec tokens, followed by an audio codec decoder to generate audio waveforms~(\Cref{fig:model-taxonomy}b). 
    
    \item \textbf{Vision-language-action models (VLAs)} are used in robotics applications to produce a robot action state from an image observation and a text instruction. VLAs often use various combinations of a text encoder, a ViT encoder, a Transformer backbone, and an action decoder~\citep{black2025pi05,wang2025alpamayo}.
    
    \item \textbf{World models} pair a video encoder with an iterative latent predictor that rolls forward over a variable horizon to understand the world. They are often used in applications such as robotic planning~\citep{assran2025vjepa2,hafner2023dreamerv3,agarwal2025cosmos,alonso2024diamond}.
    
\end{enumerate}

\subsection{Composite Models are Computation Graphs}
\label{sec:models-as-graphs}

Composite models pose three concrete challenges:

\begin{enumerate}[label=(C\arabic*), leftmargin=*, itemsep=2pt, topsep=2pt, parsep=0pt]
  \item \textbf{Architectural diversity.} As described above, multimodal models exhibit diverse architectures with multiple distinct execution paths depending on input modality. For example, in UMMs, text-to-text chat, text-to-image generation, image-to-text understanding, and image-to-image editing all use different subsets of components with distinct computation patterns.
  Many computation patterns also involve non-AR loops (e.g.,classes 1, 4, and 5 in \Cref{sec:compound-models}). %
  \item \textbf{Performant modularity.} Frameworks such as HuggingFace Transformers~\citep{wolf2019huggingface} offer broad flexibility, but often sacrifice efficiency. In contrast, specialized systems such as vLLM~\cite{kwon2023vllm} for AR text generation and VoxServe~\cite{kamahori2026voxserve} for speech generation achieve higher performance in their target domains by implementing domain-specific optimizations, but they do not generalize across modalities.

  \item \textbf{Physical topology.} Composite models use heterogeneous components that may be executed in sequence, in a pipeline, or in parallel.
  Input tensors may be batched or streamed between components across diverse links, such as intra-node NVLink or inter-node Infiniband.
  Thus, system flexibility in the physical placement and data transport between components is critical for end-to-end performance.
  
\end{enumerate}

\textbf{Why Existing Systems Fall Short.}
Existing serving systems address only a subset of these challenges.
vLLM~\citep{kwon2023vllm} and SGLang~\citep{zheng2024sglang} are highly optimized for AR text generation but treat multimodal inputs as prefill-time encoder add-ons, with no first-class support for patterns such as non-AR loops, parallel execution of nodes within a graph walk, or cross-component data streaming.
vLLM-Omni~\citep{yin2026vllmomni} and SGLang-Omni~\citep{sglangomni} extend vLLM and SGLang, respectively, to DAGs of stages glued together with explicit data transfer functions; this fits two- or three-stage thinker--talker pipelines~(\Cref{fig:model-taxonomy}b), but neither exposes loops or parallel composition of stages. Therefore, per-model glue code is needed for other patterns such as diffusion loops, fan-out in CFG, and custom policies for streaming data across components.

\textbf{Computation Graphs in \sys{}.}
Despite the architectural diversity above, we observe that every model in \Cref{sec:compound-models} has the same structural shape: a directed graph of heterogeneous \emph{components} such as encoders, decoders, and transformer backbones~(\Cref{fig:model-taxonomy}).
Intermediate tensors flow along the edges, possibly in a streaming fashion.
Each request traverses the graph over a small number of \emph{walks}.

Thus, \sys{} addresses C1 by treating multimodal inference as a graph execution, allowing individual components to be composed in arbitrary loops, chains, and parallel branches~(\Cref{sec:wg-defn}).
We avoid the abstraction tax by ``compiling'' the graph abstraction down to a high-performance serving runtime.
The runtime includes an efficient request scheduler and per-component engines that integrate state-of-the-art optimizations across modalities~(\Cref{sec:runtime}), thus addressing C2.
Finally, \sys{} addresses C3 by enabling user-defined physical placements~(\Cref{sec:capabilities}) and policies to move data from one component to another~(\Cref{sec:wg-defn}).

\begin{figure}[t]
  \centering
  \includegraphics[width=0.8\linewidth,page=5,trim={0cm 2cm 3cm 0cm},clip]{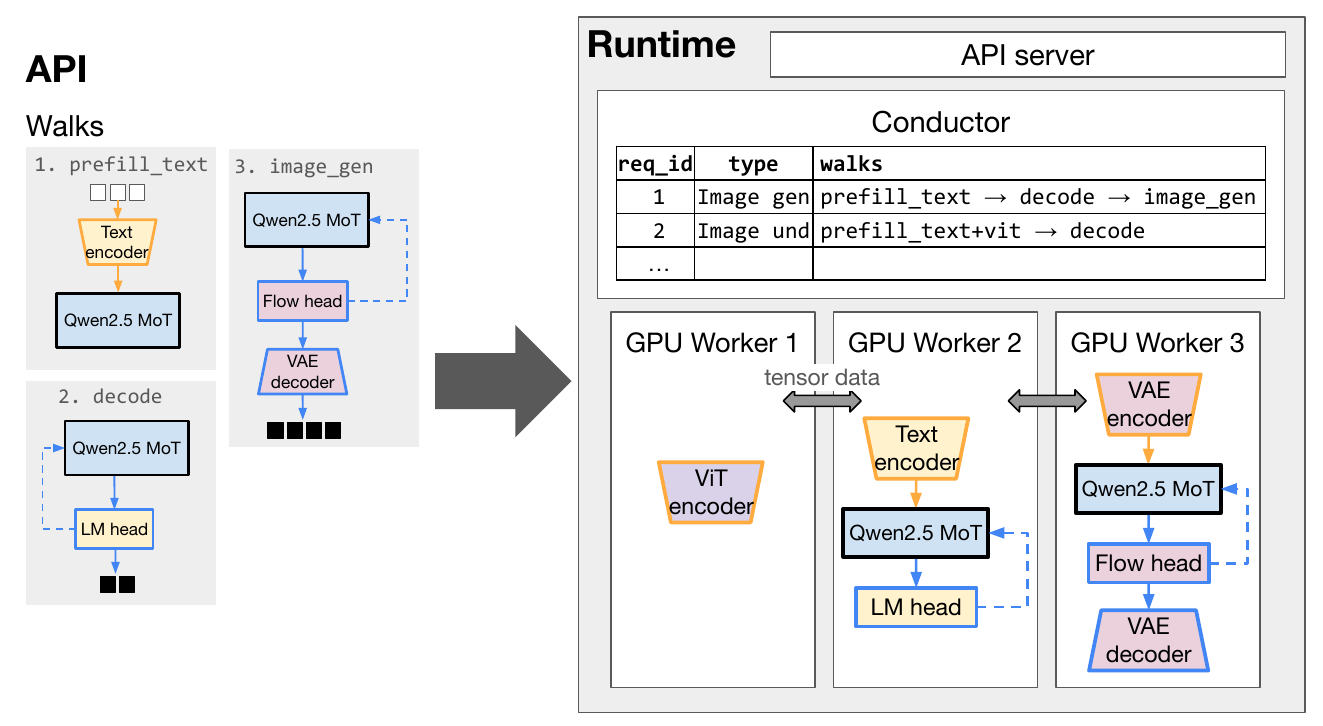}
  \caption{\sys at a glance. \textbf{Left}: The model author defines the model as a computation graph~(\Cref{fig:model-taxonomy}) and a series of graph walks.
  \textbf{Right}: The runtime~(\Cref{sec:runtime}) places the subgraphs of components on different GPU workers based on a user-specified placement.
      }
  \label{fig:system-overview}
\end{figure}

\section{The Walk Graph}
\label{sec:walkgraph}

\sys{} is built on a single contract where a model is a directed \emph{computation graph}, a request is a series of \emph{Walks} of the graph, and the runtime is the executor of the graph.  
This contract is the \textbf{Walk Graph}, a model computation graph with a finite set of named Walks.  
We define the Walk Graph in \Cref{sec:wg-defn}, enumerate the capabilities it unlocks in \Cref{sec:capabilities}, and describe the corresponding runtime in \Cref{sec:runtime}.

Table~\ref{tab:subsumption} previews how the Walk Graph relates to existing multimodal serving abstractions. 
Each prior system corresponds to a restricted subset of the Walk Graph.

\begin{table}[t]
\centering
\small
\caption{Comparing the Walk Graph against existing multimodal serving
abstractions~\citep{yin2026vllmomni,sglangomni}.
}
\label{tab:subsumption}
\setlength{\tabcolsep}{5pt}
\renewcommand{\arraystretch}{1.2}
\begin{tabular}{@{}p{3.5cm} c c c@{}}
\toprule
& \textbf{vLLM-Omni~\citep{yin2026vllmomni}} & \textbf{SGLang-Omni~\citep{sglangomni}} & \textbf{\sys{}} (ours) \\
\midrule
Graph node
  & Engine-instance stage
  & Worker-pool stage
  & \textbf{Model component} \\
Composition primitives
  & Flat DAG
  & Flat DAG
  & \textbf{Seq. / Par. / Loop / Stream} \\
Execution paths per model
  & Prefill, decode
  & Prefill, decode
  & Flexible \\
Loops
  & Within a stage
  & Within a stage
  & Across any subgraph \\
Placement granularity
  & Stage
  & Stage
  & Component, w/ optional Walk \\
\bottomrule
\end{tabular}
\end{table}

\subsection{API}
\label{sec:wg-defn}

A model is declared as a tuple $(G, W)$ where $G = (N, E)$ is a directed
\emph{computation graph} of nodes and edges, and $W = \{w_1, \dots, w_n\}$ is a finite set of named
\emph{Walks}. Each Walk is a labeled subgraph of $G$ corresponding to one
phase of model behavior, e.g., \texttt{prefill\_text} is the label for the Walk that prefills text tokens in the input prompt in \Cref{fig:system-overview}. Each request is a series of Walks, e.g., image understanding is a series of [\texttt{prefill\_text} $\to$ \texttt{prefill\_vit} $\to$ \texttt{decode}].
The model author provides a per-model state machine that determines each request's next Walk based on the request's modalities and the
outputs of the current Walk.
The model author only provides $(G, W)$ with the state machine, and the execution of requests is the job of \sys{'s} runtime.

\paragraph{Four composable primitives.}
\label{sec:wg-primitives}
The computation graph $G$ and the Walks $w_i$ are built from two types: \texttt{GraphNode}, representing a unit of computation that fires
when its required inputs arrive, and \texttt{GraphEdge}, representing a tensor flowing from one node to another. Nodes are composed into a computation (sub)graph using the four primitives (full description in Appendix~\ref{sec:primitives-full}):
\begin{itemize}\itemsep0pt\parsep0pt
  \item \texttt{Sequential}: a chain of subgraphs where the outputs of one feed into the next.
  \item \texttt{Parallel}: a fan-out of children that may execute concurrently.
  \item \texttt{Loop}: bounded iteration with per-iteration and accumulated output channels.
  \item \texttt{DynamicLoop}: \texttt{Loop} with per-request early-exit, such as for end-of-sequence (EOS) in AR models or rollout horizon in world models.
\end{itemize}

\paragraph{Streaming edges and chunk policies.}
In streaming-output models (e.g., real-time speech generation), the
producer emits one tensor at a time and the consumer must accumulate a ``chunk'' of tensors before firing.  We denote this with
\texttt{StreamingGraphEdge}, parameterized by a \texttt{ChunkPolicy}
that decides when the consumer has accumulated enough input to fire.
We observe that three policies suffice for the models in our evaluation:
(i) \texttt{FixedChunkPolicy($K$)} fires every $K$ items;
(ii) \texttt{SlidingWindowChunkPolicy($W$,$S$)} fires once the buffer
holds $W$ items and advances by $S$;
(iii) \texttt{LeftContextChunkPolicy($C$,$L$)} prepends $L$ frames of
left context to each $C$-frame chunk. The policy interface is extensible: new policies can be added without changes to the system.

\paragraph{Example: BAGEL.}
We make the abstraction concrete with BAGEL~\citep{deng2025bagel}~(\Cref{fig:model-taxonomy}a), a
unified multimodal model that handles image understanding, image
generation, and image editing with one Mixture-of-Transformers~\citep{liang2024mot}
backbone.
Appendix~\ref{sec:wg-other-models} shows examples of other models.

\begin{lstlisting}[style=walkgraph, caption={Simplified non-CFG Walk for image generation in BAGEL.}, label={fig:bagel-walkgraph}]
image_gen = Sequential([
    Loop(section=GraphNode(name="LLM", input_ids={"latents", "time_index"},
                             outputs=[GraphEdge(next_node="LLM", name="latents"),
                             GraphEdge(next_node="LLM", name="time_index")]),
         n_iters=49, outputs=[GraphEdge(next_node="vae_decoder", name="latents")]),
    GraphNode(name="vae_decoder", input_ids={"latents"},
               outputs=[GraphEdge(next_node=EMIT_TO_CLIENT, name="image_output")])])
\end{lstlisting}
BAGEL's computation graph consists of seven nodes (\texttt{vit\_encoder},
\texttt{vae\_encoder}, \texttt{LLM}, \texttt{LLM\_cfg\_text},
\texttt{LLM\_cfg\_img}, \texttt{combine\_cfg}, \texttt{vae\_decoder})
and six Walks across those nodes.
The \texttt{image\_gen} Walk is a \texttt{Sequential} combination that chains a 49-iteration \texttt{Loop} into a terminal \texttt{vae\_decoder}. Inside the \texttt{Loop} is a
\texttt{Sequential} that runs (i) a \texttt{Parallel} region containing the three CFG
branches (\texttt{LLM}, \texttt{LLM\_cfg\_text},
\texttt{LLM\_cfg\_img})
and (ii) \texttt{combine\_cfg} that applies the CFG formula
and an Euler step and then loops the updated latents back to all three
branches. After the final iteration, the resulting latents flow into the
\texttt{vae\_decoder}, which emits the decoded image to the client.
For simplicity, the listing above shows a non-CFG version of this Walk, in
which a single \texttt{LLM} node is iterated instead of three. The full CFG Walk is shown in \Cref{sec:cfg-example} of the Appendix.

\subsection{What the Walk Graph Unlocks}
\label{sec:capabilities}
\label{sec:wg-bagel}

The key contribution of the Walk Graph is that it decouples the composite model architecture from the system runtime. We enumerate the capabilities afforded by this design.

\paragraph{Modality-aware scheduling.}
The Walk abstraction allows the runtime scheduler to execute requests efficiently and in a model-agnostic manner.
The scheduler only needs to track each request's type, and its currently executing \texttt{GraphNode} and Walk~(\Cref{fig:system-overview}).
Once one Walk has finished, it then uses the state machine provided by the model author to select the next Walk.
A key benefit is that the scheduler by construction executes the minimum components needed to complete each request, rather than forcing all requests to execute all components of the model.

\paragraph{Flexible parallelism.}
Authors directly capture parallelism within their model using the graph composition primitives, e.g., using \texttt{Parallel} for CFG in BAGEL~(\Cref{sec:wg-defn}).
Other examples include $K$-way model-predictive-control for world models \cite{assran2025vjepa2}, and multi-branch sampling for AR models \cite{orpheus2025}.
The system runtime is agnostic to the specific model architecture and supports parallelism uniformly across them.
By contrast, vLLM-Omni tightly couples modality-specific features such as CFG to the system runtime by adding glue code to expand a user request into multiple branches.

\sys{} further exposes tensor parallelism (TP) as a graph-node-level abstraction.
To shard a node, an author replaces its \texttt{Linear}, \texttt{MLP}, and \texttt{Attention} layers with globally-provided sharded counterparts and declares the node's TP degree in the configuration \texttt{yaml}.
Everything else, e.g., scheduling,
synchronization, tensor transport between nodes of differing TP world sizes, and KV-cache transfer, is handled by the system runtime.

\paragraph{Flexible placement.}
The model deployer may specify a placement mapping from \texttt{GraphNode} to GPU rank(s).
This decoupled API allows for flexible placements without needing to modify the model or runtime code.
Optionally, the user can specify different GPU ranks for the same logical graph node across different Walks.
This API can be used to express common patterns important for maximizing hardware utilization, such as data-parallelism by specifying multiple ranks or prefill-decode disaggregation~\cite{zhong2024distserve} by specifying different ranks for prefill vs.~decode Walks.
In addition, it enables independent scaling of different model components, e.g., placing encoders and decoders on many small GPUs and LLM backbones on few large GPUs.

Furthermore, this approach allows for transparent resource sharing across concurrent requests, even ones of different types or Walks.
For example, in BAGEL the \texttt{LLM} node appears in all Walks~(\Cref{sec:wg-defn} and \Cref{fig:system-overview}): if the user specifies the same GPU rank(s) for the node in all Walks, then the system automatically multiplexes the same physical \texttt{LLM} replica(s) across requests from all Walks.

\paragraph{Loop optimizations.}
\texttt{Loop} and \texttt{DynamicLoop} provide first-class loop support and can be used to express iterative patterns including AR decoding until EOS, fixed-step diffusion, and per-request rollout horizon in world models.
These semantics are important for enabling portability of system optimizations across different model architectures.
In contrast, stage-DAG abstractions in vLLM-Omni and SGLang-Omni cannot have cycles; therefore, any loops must remain internal to a stage.

The \texttt{Loop} abstraction allows performance features such as CUDA graphs and continuous batching~\citep{yu2022orca} to be agnostic to the presence of loops.
Furthermore, it enables scheduling of loop iterations as if they were any other component.
For instance, in the BAGEL model, the \sys{} scheduler can seamlessly interleave flow steps and autoregressive decoding steps that use the same \texttt{LLM} node.

\paragraph{Flexible chunk policies.}
\texttt{StreamingGraphEdge(policy)} allows capture of arbitrary producer-consumer patterns within a model architecture. Three reusable \texttt{ChunkPolicy} types
(\S\ref{sec:wg-primitives}) cover every streaming connection in our
evaluation.
For example, Qwen3-Omni uses a \texttt{FixedChunkPolicy} with chunk size 1 for the Thinker$\to$Talker connection; this means that the Talker should consume each Thinker output as soon as it arrives.
The same model uses \texttt{LeftContextChunkPolicy} for Talker$\to$Code2Wav for causal smoothing of the audio output.
Critically, the runtime is agnostic to the chunk policy used; the same infrastructure is used for a range of streaming patterns.

\subsection{Runtime}
\label{sec:runtime}

The runtime executes requests for the expressed Walk Graph. An HTTP server accepts
requests; a \emph{Conductor} (one per server) maintains per-request
Walk state and dispatches work to \emph{Workers} via ZeroMQ~\citep{hintjens2013zeromq};
Workers (single-process; one per GPU rank) execute the local subgraph~(\Cref{fig:system-overview}), routing tensors directly to downstream workers.
Cross-rank graph edges are inter-process tensor transfers,  with streaming edges
instantiating a per-request input buffer at the consumer.
The data plane is pluggable and supports shared memory, as well as RDMA and TCP via Mooncake~\citep{qin2025mooncake}.

Each \texttt{GraphNode} is executed by an \emph{engine}, an inference instance selected by the model author based on the node’s component type.
There are currently two engines:  \texttt{KVCacheEngine}, a modality-agnostic transformer engine with \texttt{FlashInfer}-based paged-attention KV-cache state and a cuda-graph-compatible sampling plugin.
Stateless nodes use simpler execution paths via \texttt{StatelessEngine}.
Both engines support continuous batching, CUDA-graph replay, and \texttt{torch.compile}.
Each worker can host multiple engines and runs a local scheduler to drive the engines with a round-robin execution policy.

To overlap CPU scheduling with GPU execution, each worker asynchronously schedules batch $N+1$ while batch $N$ is still in flight.
The next batch is scheduled by alternating between the following: (1) traversing the Walk Graph with the outputs of the current batch $N$, and determining what nodes will be ready once the current batch finishes, and (2) scheduling any unrelated batches that are ready (to avoid head-of-line blocking).
The heaviest overhead is often constructing the FlashInfer attention plan; we double-buffer the attention plan and asynchronously construct the next attention plan in a separate thread and CUDA stream.
For speculation across \texttt{DynamicLoop} iterations, termination checks are deferred to the next iteration, so each termination costs at most one wasted step.\footnote{For models where this wasted step is inadmissible, speculative scheduling can be disabled on a per-node level.}

\definecolor{Mstar}{RGB}{31,119,180}     
\definecolor{Baseline}{RGB}{255,127,14}  

\section{Evaluation}
\label{sec:eval}
\owh{}

We evaluate \sys{} on BAGEL-7B~\citep{deng2025bagel}, Qwen3-Omni-30B-A3B~\citep{xu2025qwen3omni}, Orpheus-3B~\citep{orpheus2025}, and V-JEPA~2 vitg-AC~\cite{assran2025vjepa2}. As baselines, we use vLLM-Omni for supported models (BAGEL and Qwen3-Omni), SGLang-Omni for Qwen3-Omni, and modality-specific baselines for the remainder: VoxServe~\citep{kamahori2026voxserve} for Orpheus and Meta’s native \texttt{vjepa2} implementation for V-JEPA~2.
Since $\pi_{0.5}$ and V-JEPA~2-AC are both robotic planning models, are unsupported by HuggingFace Transformers and serving frameworks such as vLLM-Omni, and only provide native repositories, we benchmark only V-JEPA~2-AC.
All experiments were run on either a single 4$\times$H100 node or a single 8$\times$H200 node; configurations are reported inline.

\textbf{Metrics.} When measuring text outputs, we report time-to-first-token (TTFT) and throughput. For image generation, we report end-to-end (E2E) latency.
For audio, we report the per-request \textbf{real-time factor (RTF)}, the ratio of processing wall time to generated audio duration.
Lower is better and $<$$1$ means streaming is feasible.
Each configuration uses 5-10 warmup requests followed by 10--160 timed requests (at least $5\times$ the maximum concurrency).
We report p50 (solid bars) and p95 (hatched extensions) where appropriate.

\subsection{BAGEL: \sys{} vs vLLM-Omni}
\label{sec:eval-bagel}

We evaluate BAGEL on text-to-image (T2I), image editing (I2I), and image-to-text understanding (I2T) using inputs from VBench~\cite{huang2024vbench} for generation tasks and Food101~\cite{bossard14food101} for understanding.
Both systems use the BAGEL-7B checkpoint, generating images ar $1024\times 1024$ resolution with a 50-step flow schedule.
For I2I, both \sys{} and vLLM-Omni generate images of the same aspect ratio as the input image (scaled such that the long edge is dimension 1024).

For T2I/I2I, we benchmark two configurations of vLLM-Omni: the default configuration, which has a ``Thinker'' and diffusion transformer (DiT) stage (essentially replicating the BAGEL transformer), and a single-stage configuration. For I2T, the default configuration outperforms the single-stage pipeline, so we benchmark the default configuration, with the maximum number of sequences set to largest batch size tested.

For T2I/I2I we use 3 H100s with classifier-free-guidance (CFG) parallelism (one rank per CFG branch). \sys{} runs the three branches in parallel via the \texttt{Parallel} primitive~(\Cref{sec:wg-defn}); vLLM-Omni uses a specialized CFG parallel plugin that ues \texttt{torch.distributed}.

For I2T we use 1 H100, since CFG only applies to image generation.
To ensure exact parity in output token count, we ignore EOS in both vLLM-Omni and \sys{}, instead generating until a benchmark-determined sequence length.
All I2T results are averaged across three benchmark runs.

\begin{figure}[t]
  \centering
    \centering
    \includegraphics[width=0.65\linewidth]{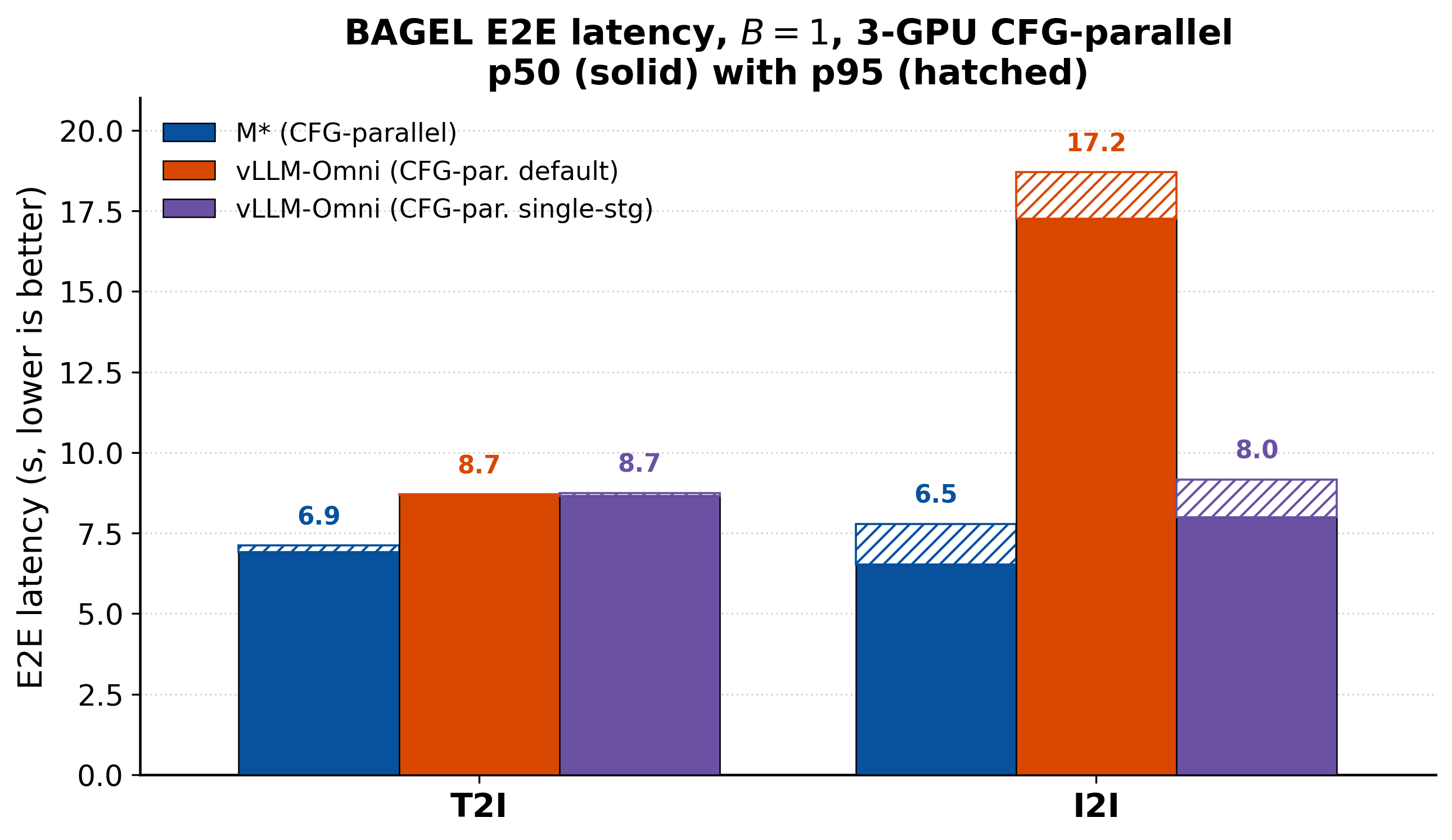}
    \caption{\small\textbf{BAGEL T2I/I2I E2E latency, $B{=}1$, 3-GPU CFG-parallel.}
    }
    \label{fig:bagel-e2e-workloads}
\end{figure}

\textbf{Image generation (T2I, I2I).} On 3-GPU CFG-parallel at $B{=}1$, \sys{} improves on single-stage vLLM-Omni's p50 end-to-end latency by $1.25\times$ on T2I and $1.22\times$ on I2I~(Fig.~\ref{fig:bagel-e2e-workloads}).
For the default configuration, which involves an expensive KV cache transfer between the Thinker and DiT, our advantage on I2I grows to $2.64\times$.
The p95 advantage is similar.
We also see performance gains in the single-GPU (i.e., no CFG parallelism) case, plots for which can be found in \Cref{sec:deferred-experiments}.

This improvement is primarily due to \sys{}'s KV cache management abstractions: we represent the three CFG contexts as three labels over a single paged KV pool, as opposed to vLLM-Omni's dense \texttt{NaiveCache} per CFG context.
Each denoise step can then apply paged attention, reading page tables in place; vLLM-Omni concatenated key and value tensors at every layer and for every step.
The label is a general cache-key axis, as such, it inherits paging, offload, by-reference transfer, and continuous batching, whereas the dense \texttt{NaiveCache} is model-specific without immediate access to such optimizations.

\begin{figure}[t]
    \centering
    \includegraphics[width=0.99\linewidth]{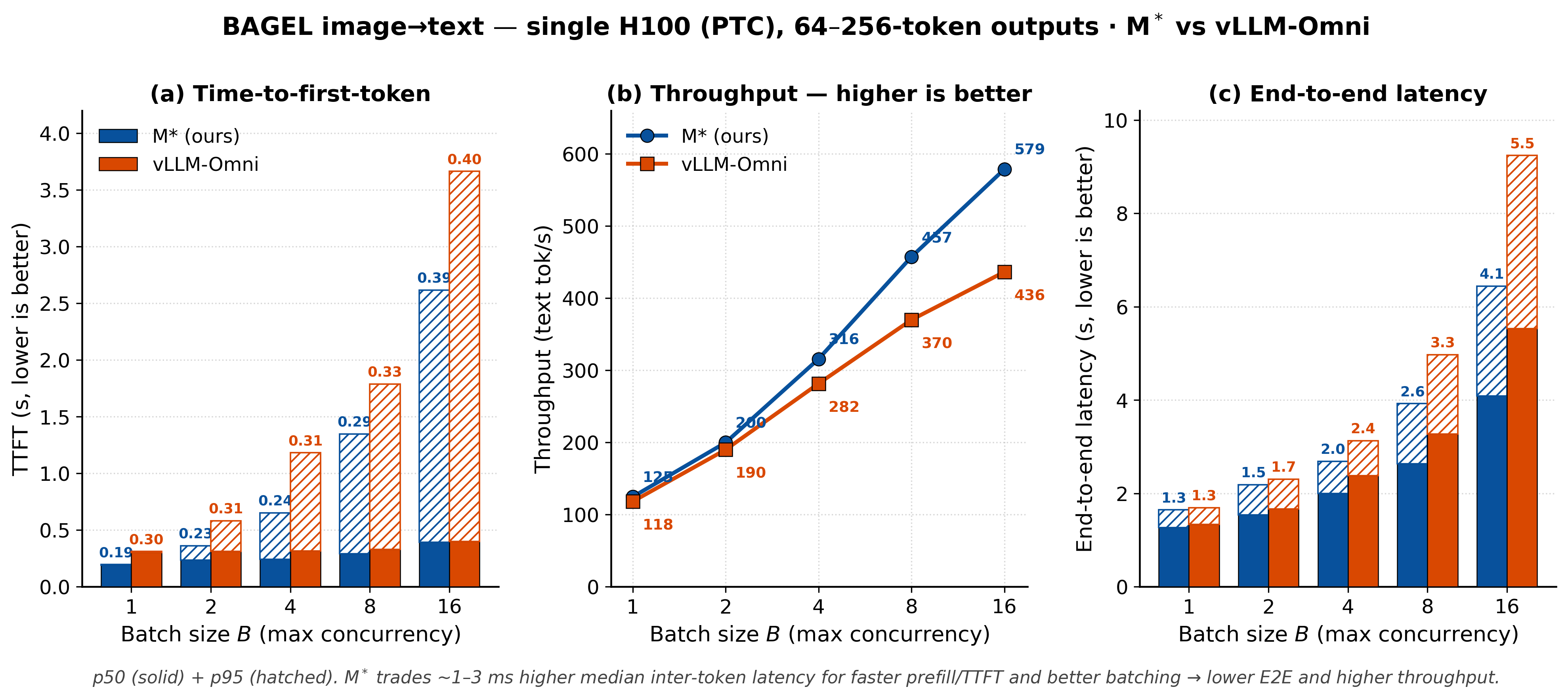}
    \caption{\small\textbf{BAGEL I2T, single H100, $B \in \{1, 2, 4, \dots, 16\}$.}, \texttt{ignore\_eos} with output lengths between 64 and 256. (a) TTFT (log $y$). (b) Throughput (req/s). (c) End-to-end request latency.
    }
    \label{fig:bagel-i2t-1gpu}
\end{figure}

\textbf{Image understanding (I2T).} 
For output token lengths distributed uniformly between 64 and 256, we achieve comparable throughput to vLLM-Omni at low batch sizes, with our advantage increasing to 32.7\% by $B{=}16$ (\Cref{fig:bagel-i2t-1gpu}).
This is also reflected in the per-request E2E latency (25.5\% improvement at $B{=}16$).
\sys{} achieves consistently lower p50 TTFT across batch sizes, ranging from 33\% at $B{=}1$ to 14\% at $B{=}16$, while maintaining a tighter tail: \sys{} p95 TTFT is 28\% lower than that of vLLM-Omni at $B{=}16$, despite lower p50 gains.
Figures~\ref{fig:bagel-i2t-1gpu-short} and \ref{fig:bagel-i2t-1gpu-long} in the Appendix (varying the output token length distribution) show a similar story , with our advantage most prominent for shorter-decode workloads.

\textbf{\sys{} enables one config for all modalities.}
In this section, we benchmark two configurations of vLLM-Omni, and neither performs well across all three workloads at once: the default configuration performs well for T2I and I2T but poorly for I2I, whereas the single-stage configuration performs well for T2I and I2I but poorly for I2T.\footnote{Specifically, it achieves a throughput of 41 tokens/sec on batch size 1 (with \texttt{enforce\_eager} manually set to \texttt{false} in the config file, and \texttt{max\_num\_seqs} increased to \texttt{16}), which is half as fast as the default config. It also appears to not support continuous batching or streaming of tokens to the API server.}
The default config suffers on I2I because the Thinker and DiT are separate stages, requiring an expensive KV-cache transfer between prefill and the flow loop; the single-stage configuration collapses them into one process and foregoes that transfer, but no longer runs I2T on vLLM's optimized AR engine---losing continuous batching and token streaming.
\sys{}, by contrast, serves optimized T2I, I2I, \textit{and} I2T with the same configuration, while also enabling PD disaggregation, encoder disaggregation, and tensor parallelism with minor (mainly config-level) changes.

\subsection{Qwen3-Omni: \sys{} vs vLLM-Omni and SGLang-Omni}
\label{sec:eval-qwen3-omni}

\begin{figure}[t]
  \centering
  \includegraphics[width=0.8\linewidth]{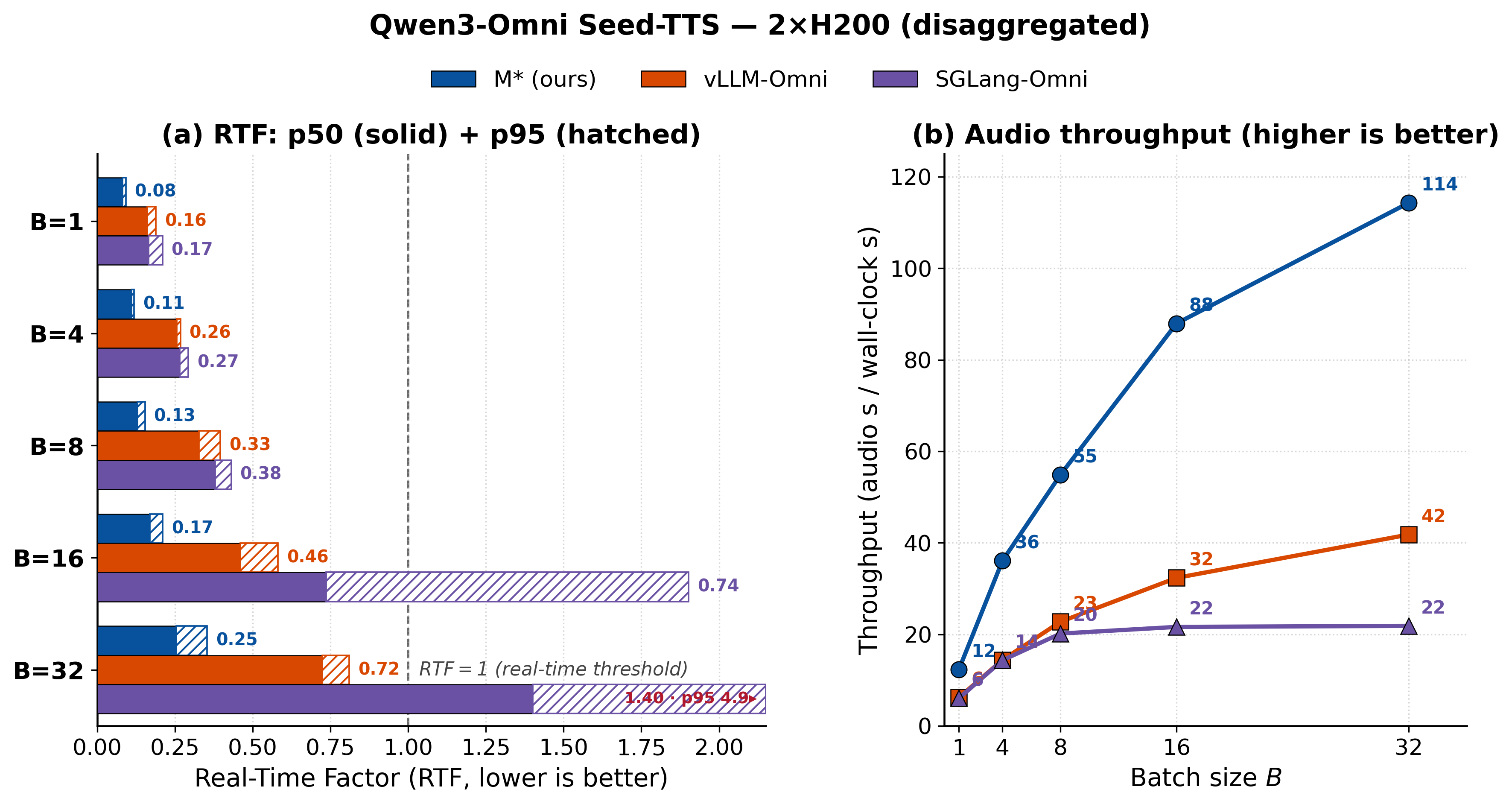}
  \caption{\small\textbf{Qwen3-Omni Seed-TTS, 2-GPU.} (a) RTF (lower is better).
  (b) Audio throughput (higher is better).
  }
  \label{fig:qwen3-omni-eval}
\end{figure}

\begin{figure}[t]
  \centering
  \includegraphics[width=0.8\linewidth]{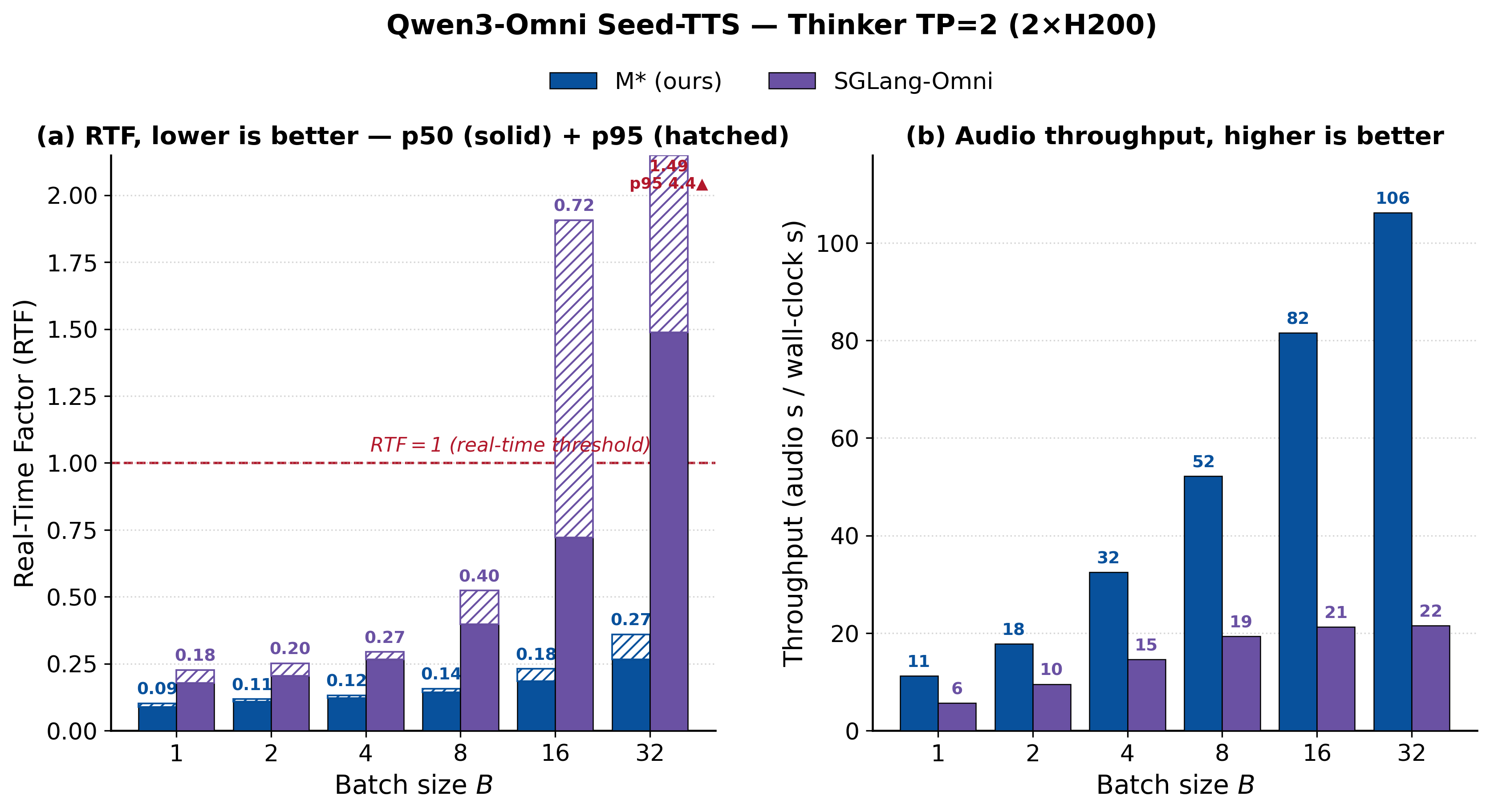}
  \caption{\small\textbf{Qwen3-Omni Seed-TTS, Thinker TP 2 (3-GPU.} (a) RTF (lower is better).
  (b) Audio throughput (higher is better).
  }
  \label{fig:qwen3-omni-tp2-eval}
\end{figure}

We run Qwen3-Omni on the Seed-TTS~\cite{anastassiou2024seedtts} text-to-speech benchmark on 2 H200s, with the model disaggregated as Thinker on rank~0 and Talker$+$Code2Wav on rank~1 (\Cref{fig:qwen3-omni-eval}). \sys{} significantly outperforms vLLM-Omni and SGLang-Omni's RTF across batch sizes.
At $B{=}16$, \sys{} delivers $2.7\times$ and $4.0\times$ higher throughput than vLLM-Omni and SGLang-Omni, respectively.
We also apply degree-2 tensor parallelism to the Thinker (overall using 3 H200s), and compare against SGLang-Omni.\footnote{We were unable to get vLLM-Omni's tensor-parallel Thinker to work.}
As shown in \Cref{fig:qwen3-omni-tp2-eval}, our RTF and throughput advantage remain consistent to the non-TP results in \Cref{fig:qwen3-omni-eval}.


The RTF and throughput improvement is largely due to the \sys{}'s flexibility and modularity.
As CUDA graph capture in \sys{} is defined on a per-submodule level with customizable inputs and outputs, the  entire Talker submodule, including the multi-token predictor loop, is able to run as a single CUDA graph.
vLLM-Omni, on the other hand, has CUDA graphs explicitly disabled for the Code Predictor.
In addition, our system places a co-located Talker and Code2Wav on the same Worker process (via our one-to-one Worker-to-GPU mapping), eliminating the need for inter-process communication of Talker codes to the Code2Wav.
Both vLLM-Omni and SGLang-Omni require separate processes for each stage.
The \texttt{StreamBuffer} abstraction allows clean support for both colocated and disaggregated streaming of data between graph nodes.

\subsection{Orpheus: \sys{} vs VoxServe}
\label{sec:eval-orpheus}

\begin{figure}[t]
  \centering
  \includegraphics[width=0.75\linewidth]{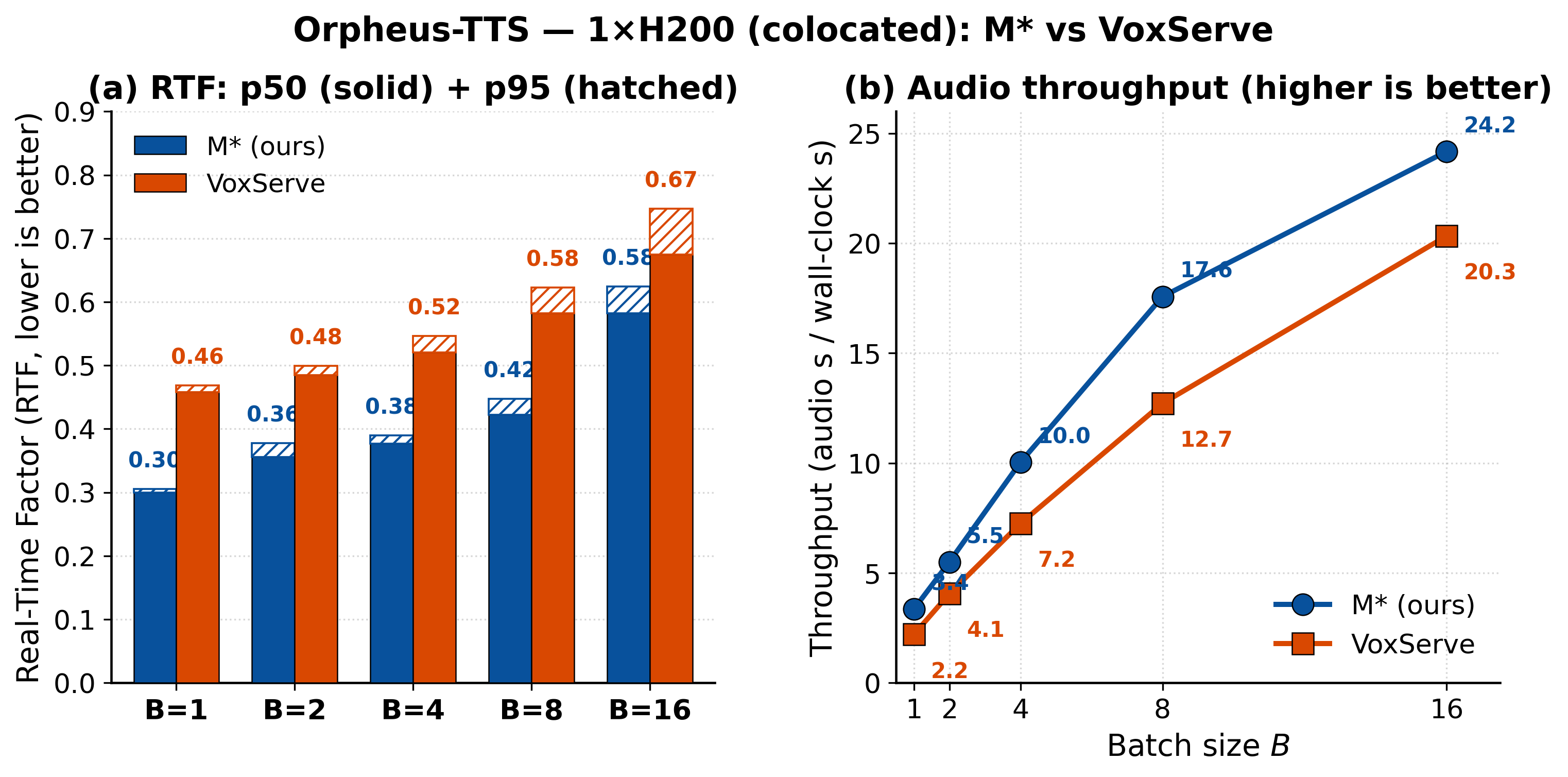}
  \caption{\small\textbf{Orpheus TTS, single H200.} (a) RTF (lower is better). (b) Audio throughput (higher is better).
  }
  \label{fig:orpheus-eval}
\end{figure}

We measure Orpheus-3B performance on a single H200 against VoxServe~\citep{kamahori2026voxserve} on $B$=$\{1,2,4,8,16\}$ and Seed-TTS~(Fig.~\ref{fig:orpheus-eval}), averaging results over 5 trials.
For the \sys{} implementation, we have an LLM with a \texttt{StreamingEdge} into a SNAC audio decoder.
\sys{} performance is overall better: \sys{} delivers $13.6\%$ lower p50 RTF than VoxServe at $B{=}16$, with throughput improvements ranging from $20\%$ ($B{=}8$) to $52\%$ ($B{=}1$). We attribute these gains to the following components of \sys{}: Authors inherit model-level improvements for free, e.g., fused projections provided by \sys{}'s \texttt{Attention} layers and cuda-graph-compatible sampling, and the Walk Graph abstraction enables the speculative \texttt{FlashInfer} planning described in \S\ref{sec:runtime} for arbitrary multimodal requestss.

\subsection{V-JEPA 2: Rollout for Robotic Planning}
\label{sec:eval-vjepa2}

V-JEPA~2 is a world model that supports action-conditioned (AC) rollout: each step autoregressively predicts the next video frame conditioned on an action and previous states. We compare against the Meta's \texttt{vjepa2} implementation on 1 H100 with $B{=}1$ at rollout horizons $H$=$\{4, 15, 30\}$, using inputs from the first 50 episodes of the DROID dataset~\cite{khazatsky2024droid}. The baseline runs a hand-written Python autoregressive loop without KV caching, forcing costly prefill over a growing sequence at every iteration.

\begin{figure}[tbp]
  \centering
    \includegraphics[width=0.70\linewidth]{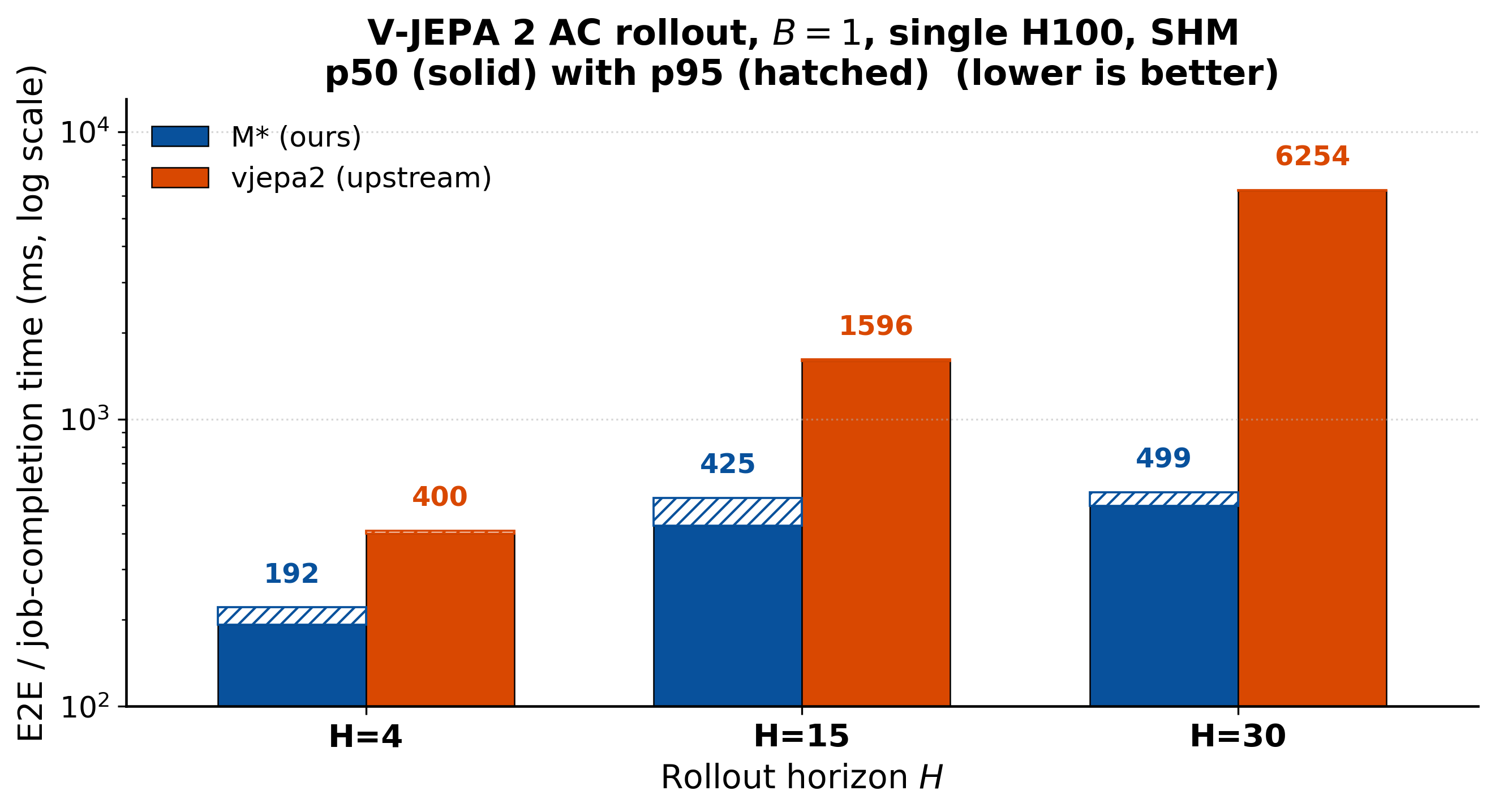}
    \caption{\small\textbf{V-JEPA~2 AC rollout, $B{=}1$, single H100} (lower is better).
    }
    \label{fig:vjepa-rollout}
\end{figure}

Meanwhile, \sys{} encodes the rollout as a \texttt{DynamicLoop} over the \texttt{AREngine}, applying paged-attention KV caching to avoid duplicate prefills. Thus, \sys{} delivers p50 speedups of $2.08\times$ at $H{=}4$, $3.76\times$ at $H{=}15$, and $12.5\times$ at $H{=}30$~(\Cref{fig:vjepa-rollout}).

\section{Related Work}
\label{sec:related}

\textbf{Token-centric LLM serving.}
vLLM~\citep{kwon2023vllm}, SGLang~\citep{zheng2024sglang}, and Orca~\citep{yu2022orca} target autoregressive text generation with optimizations such as continuous batching, paged attention, and radix caching.
\sys{} reuses these optimizations and generalizes them to other modalities~(\S\ref{sec:runtime}).
DistServe~\citep{zhong2024distserve}, Splitwise~\citep{patel2024splitwise}, and Mooncake~\citep{qin2025mooncake} disaggregate prefill from decode to avoid inter-phase interference.
\sys{} supports this and other disaggregation options via flexible placement policies~(\S\ref{sec:capabilities}).

\textbf{Multimodal serving.}
vLLM-Omni~\citep{yin2026vllmomni} and SGLang-Omni~\citep{sglangomni} represent a model as a fixed DAG of stages, where each stage executes as a separate inference engine. 
ModServe~\citep{qiu2025modserve} and EPDServe~\citep{singh2024efficiently} support flexible disaggregation but only for the image preprocessing and encoder components.
Cornserve~\citep{ma2025cornserve} supports any-to-any models by recovering a dependency graph of Python tasks during execution via record-and-replay.
\sys{}'s Walk Graph captures computation ahead of time and at a finer granularity -- a forward pass of one component such as one denoising iteration in a DiT vs.~a full request on a DiT engine in Cornserve -- to enable compile-time optimizations across components.

\textbf{Modality-specific serving.}
Many works focus on modality-specific inference optimizations, most of which can be integrated into \sys{}.
VoxServe~\citep{kamahori2026voxserve} unifies SpeechLM serving with a streaming-aware scheduler; \sys{} supports this through streaming graph edges.
FastVideo~\citep{zhang2025fast,zhang2025vsa} introduces sparse attention techniques for video generation that can be directly integrated into \sys{}.
xDiT~\citep{fang2024xdit} proposes parallelism strategies for DiTs~\citep{fang2025pipefusion,fang2024unified,sun2024unveiling}, which future versions of \sys{} can support through flexible placement policies.
Inferix~\citep{team2025inferix} targets long-video world models with block-diffusion decoding, combining LLM-style KV-cache management with block-wise diffusion; \sys{}'s Walk Graph can express such hybrid autoregressive--diffusion pipelines as a single graph.
FlashDrive~\citep{li2026flashdrive} accelerates VLA inference for autonomous driving through temporal KV-cache reuse, speculative decoding, and adaptive flow-matching steps, all complementary to \sys{}'s graph-level scheduling.

\section{Conclusion}
\label{sec:conclusion}

\sys{} is a universal serving runtime built on the observation that composite multimodal models can be captured as walks over a dataflow graph.
By introducing the Walk Graph, a small set of unifying and composable graph primitives, \sys{} decouples the model architecture from the physical placement and execution.
We show that the resulting system can deliver performance on par with or better than state-of-the-art baselines across a range of model families.
As models are increasingly deployed in the real world, such composite multimodal models will become increasingly critical to end applications.
Thus, we expect that \sys{} will accelerate the development of the next generation of models.

{\small
\bibliographystyle{plainnat}
\bibliography{references}
\appendix
\appendix
\section*{Appendix}
\addcontentsline{toc}{section}{Appendices}

\section{Two More Walk Graphs at a Glance}
\label{sec:wg-other-models}

The same four primitives express qualitatively different model
families. 
Two examples make the range concrete.
\paragraph{Qwen3-Omni: three partitions, two streaming edges.}
Qwen3-Omni~\citep{xu2025qwen3omni} is an omni-modality LLM in a
three-partition topology: a Thinker (text-out AR LLM), a Talker
(codec-token AR LLM), and Code2Wav (audio-codec vocoder), each on its
own rank.  The Thinker streams hidden states to the Talker via a
\texttt{FixedChunkPolicy($K{=}1$)} \texttt{StreamingGraphEdge}, and
the Talker streams codec frames to Code2Wav via a
\texttt{LeftContextChunkPolicy}.  Eight Walks span the three
partitions, including separate prefill Walks per input modality and
separate prefill / last-prefill / decode Walks for the Talker.

\paragraph{V-JEPA~2: five Walks selected per request.}
V-JEPA~2~\citep{assran2025vjepa2} is a video world model whose
predictor is reused across distinct tasks.  Its action-conditioned
variant declares five Walks: a single-shot \texttt{prefill\_video};
an \texttt{encoder\_only} Walk for cross-Walk pre-encoding; a batched
\texttt{rollout} Walk built around a \texttt{DynamicLoop} with
per-request horizon $H$; a \texttt{streaming\_rollout} variant that
emits each iter's prediction immediately; and an MPC Walk that runs
the predictor with $K$ candidate action sequences in one batched
forward, scored by an \texttt{mpc\_scorer} node --- $K$-way
model-predictive control as a 3-node \texttt{Sequential}.  The same
predictor weights serve all five Walks.  The masked-predictor variant
declares the same set minus the MPC Walk.

\section{Walk Graph Primitives: Full Signatures and Semantics}
\label{sec:primitives-full}

Table~\ref{tab:primitives} gives the full signatures and semantics of
the four composable primitives summarized in
\S\ref{sec:wg-primitives}, plus the streaming edge variant
\texttt{StreamingGraphEdge}.

\begin{table}[!ht]
\centering
\small
\caption{The four composable primitives that generate a Walk Graph,
plus the streaming variant of \texttt{GraphEdge}.  \texttt{GraphNode}
and \texttt{GraphEdge} are the atomic types; the composite primitives
below close under nesting.}
\label{tab:primitives}
\setlength{\tabcolsep}{4pt}
\renewcommand{\arraystretch}{1.2}
\begin{tabular}{@{}l p{3.6cm} p{6.6cm}@{}}
\toprule
\textbf{Primitive} & \textbf{Signature} & \textbf{Semantics} \\
\midrule
\texttt{Sequential}
  & $\texttt{list[Section]} \to \texttt{Section}$
  & Chain --- outputs of one section feed the next. \\
\texttt{Parallel}
  & $\texttt{list[Section]} \to \texttt{Section}$
  & Fan-out --- children execute concurrently on (possibly distinct) ranks. \\
\texttt{Loop}
  & $\texttt{Section} \times \texttt{int} \to \texttt{Section}$
  & Bounded iteration.  Two output channels: \texttt{outputs} (cache wiped per iter --- only the last iter's tensor flows downstream) and \texttt{accumulated\_outputs} (cache persists across iters --- every iter's contribution is concatenated and emitted en bloc).  Disjoint by name. \\
\texttt{DynamicLoop}
  & $\texttt{Section} \times \texttt{int}_{\max} \times \texttt{str} \to \texttt{Section}$
  & \texttt{Loop} with per-request early-exit.  A submodule signals stop via \texttt{request\_info.register\_loop\_stop(name)}; on the next iter boundary the runtime advances to terminal outputs.  Used for EOS and rollout horizon. \\
\midrule
\texttt{StreamingGraphEdge}
  & $\texttt{str} \times \texttt{ChunkPolicy} \to \texttt{Edge}$
  & Streaming variant of \texttt{GraphEdge}.  Producer emits one tensor at a time; the consumer's \texttt{StreamBuffer} accumulates and gates dispatch via the policy. \\
\bottomrule
\end{tabular}
\end{table}

\FloatBarrier

\section{YAML Placement Listings}
\label{sec:disagg-yaml}

The two listings below ground the disaggregation patterns described
in \S\ref{sec:capabilities} (capabilities~3 and~4).
 \begin{minipage}{\textwidth}
\begin{lstlisting}[style=walkgraph,label=lst:qwen3omni-pd,caption={Per-Walk placement: Qwen3-Omni Thinker prefill/decode disaggregation.}]
node_groups:
  - node_names: [Thinker]
    graph_walks: [prefill_text, prefill_audio, prefill_vision]
    ranks: [0]
  - node_names: [Thinker]
    graph_walks: [thinker_decode]
    ranks: [1]
\end{lstlisting}
\end{minipage}
 \begin{minipage}{\textwidth}
\begin{lstlisting}[style=walkgraph,label=lst:bagel-cfg,caption={Declarative parallelism: BAGEL CFG-parallel placement (one rank per branch).}]
node_groups:
  - node_names: [LLM]
    ranks: [0]
  - node_names: [LLM_cfg_text]
    ranks: [1]
  - node_names: [LLM_cfg_img]
    ranks: [2]
  - node_names: [combine_cfg, vae_decoder]
    ranks: [0]
\end{lstlisting}
\end{minipage}

\section{Detailed Subsumption Table}
\label{sec:subsumption-full}

Table~\ref{tab:subsumption-full} expands the simplified comparison in
\S\ref{sec:walkgraph} (Table~\ref{tab:subsumption}) along nine axes,
each tracing back to a primitive or capability introduced in
\S\ref{sec:walkgraph}.

\begin{table}[!htbp]
\centering
\scriptsize
\caption{The Walk Graph compared with existing multimodal serving
abstractions.  vLLM-Omni and SGLang-Omni implement stage graphs at
\emph{engine-instance} granularity; VoxServe specializes a single
\texttt{Model} interface for SpeechLMs.  Each prior system corresponds
to a restriction of the Walk Graph (see \S\ref{sec:walkgraph}).}
\label{tab:subsumption-full}
\setlength{\tabcolsep}{4pt}
\renewcommand{\arraystretch}{1.15}
\begin{tabular}{@{}p{2.6cm} p{2.7cm} p{2.7cm} p{2.4cm} p{2.7cm}@{}}
\toprule
& \textbf{vLLM-Omni} & \textbf{SGLang-Omni} & \textbf{VoxServe} & \textbf{\sys{} (ours)} \\
\midrule
Graph granularity
  & Stage = engine instance
  & Stage = worker pool
  & Single \texttt{Model} class
  & \textbf{Node = one forward pass} \\
Walks per model
  & One pipeline (frozen) per \texttt{model\_type}
  & One \texttt{PipelineConfig} per variant
  & Single forward path
  & \textbf{First-class}; e.g.\ BAGEL: 6 walks share one node set \\
Cross-walk node sharing
  & ---
  & ---
  & N/A
  & Yes; \texttt{LLM} node is in every BAGEL walk \\
Typed primitives
  & Flat DAG + binary \texttt{async\_chunk} flag
  & Flat DAG + sources / aggregators
  & None
  & \texttt{Sequential / Parallel / Loop / DynamicLoop / StreamingGraphEdge} \\
Iteration as graph structure
  & Hidden in engine / model
  & Hidden in executor (e.g.\ \texttt{chunked\_decode} \texttt{while}-loop)
  & Hidden in scheduler's outer loop
  & Graph-level \texttt{Loop} / \texttt{DynamicLoop} \\
Iter loops $\Rightarrow$ CUDA graphs
  & No: \texttt{enforce\_eager: true} on Code2Wav \& BAGEL DiT \emph{(verbatim: ``cudagraph-incompatible'')}
  & Per-stage; codec executors are eager Python loops
  & Audio-only
  & Yes --- per-iter \texttt{forward} is shape-static \\
Streaming policies
  & Single \texttt{async\_chunk} toggle
  & Generic \texttt{StreamQueue} transport; chunking is per-model
  & Per-model \texttt{detokenize\_interval} (audio only)
  & Three reusable \texttt{ChunkPolicy} types cover taxonomy \\
Placement granularity
  & Per-stage \texttt{devices: str}
  & Per-stage \texttt{gpu\_placement: dict}
  & AR / detokenizer split (fixed cuda:0 / cuda:1)
  & Per-(node, walk) \\
Multimodal models supported
  & Qwen2.5/3-Omni, BAGEL, MiMo-Audio, + DiTs via Diffusers
  & Qwen3-Omni, Ming-Omni, FishAudio S2-Pro, Voxtral-TTS
  & 8 SpeechLM families (TTS + STS only)
  & BAGEL, Qwen3-Omni, V-JEPA~2 (masked + AC), $\pi_{0.5}$, Orpheus \\
\bottomrule
\end{tabular}
\end{table}

\section{Deferred Experimental Results}
\label{sec:deferred-experiments}
Figure \ref{fig:bagel-t2i-i2i-1gpu} compares vLLM-Omni with \sys{} for T2I and I2I job completion time, on a single H100 (no CFG-parallelism).
\begin{figure}[!htbp]
    \centering
    \includegraphics[width=0.85\linewidth]{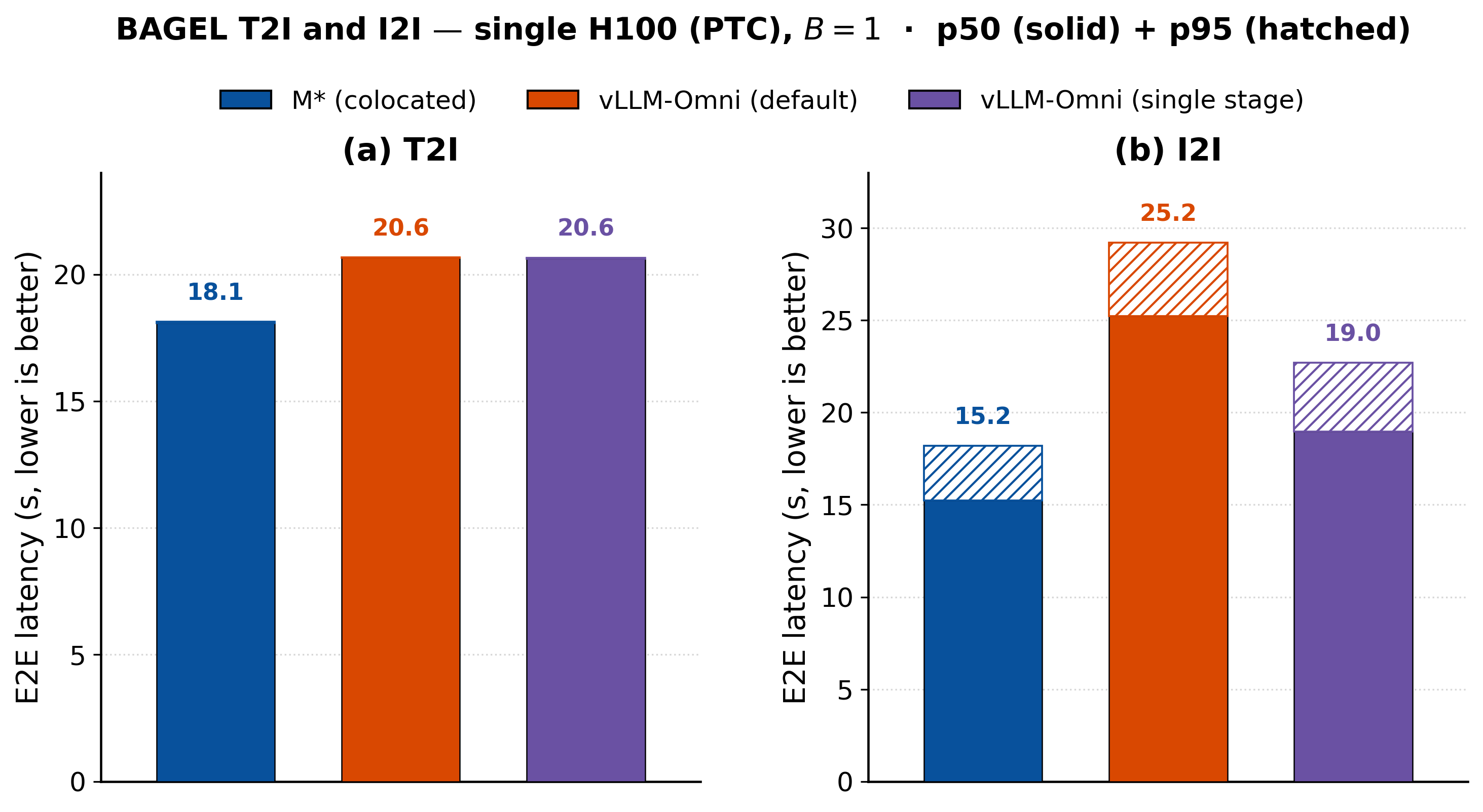}
    \caption{\textbf{BAGEL T2I and I2I, single H100, $B{=}1$.} \sys{} wins on both; the gap widens for I2I.}
    \label{fig:bagel-t2i-i2i-1gpu}
\end{figure}

\sys{} improves on vLLM-Omni's T2I p50 latency by $1.13\times$.
For I2I, \sys{} also improves on default config vLLM-Omni's I2I p50 latency by 
$1.66\times$ and $1.25\times$ for vLLM Omni's single-stage config (\Cref{fig:bagel-t2i-i2i-1gpu}).

\begin{figure}[!htbp]
    \centering
    \includegraphics[width=0.99\linewidth]{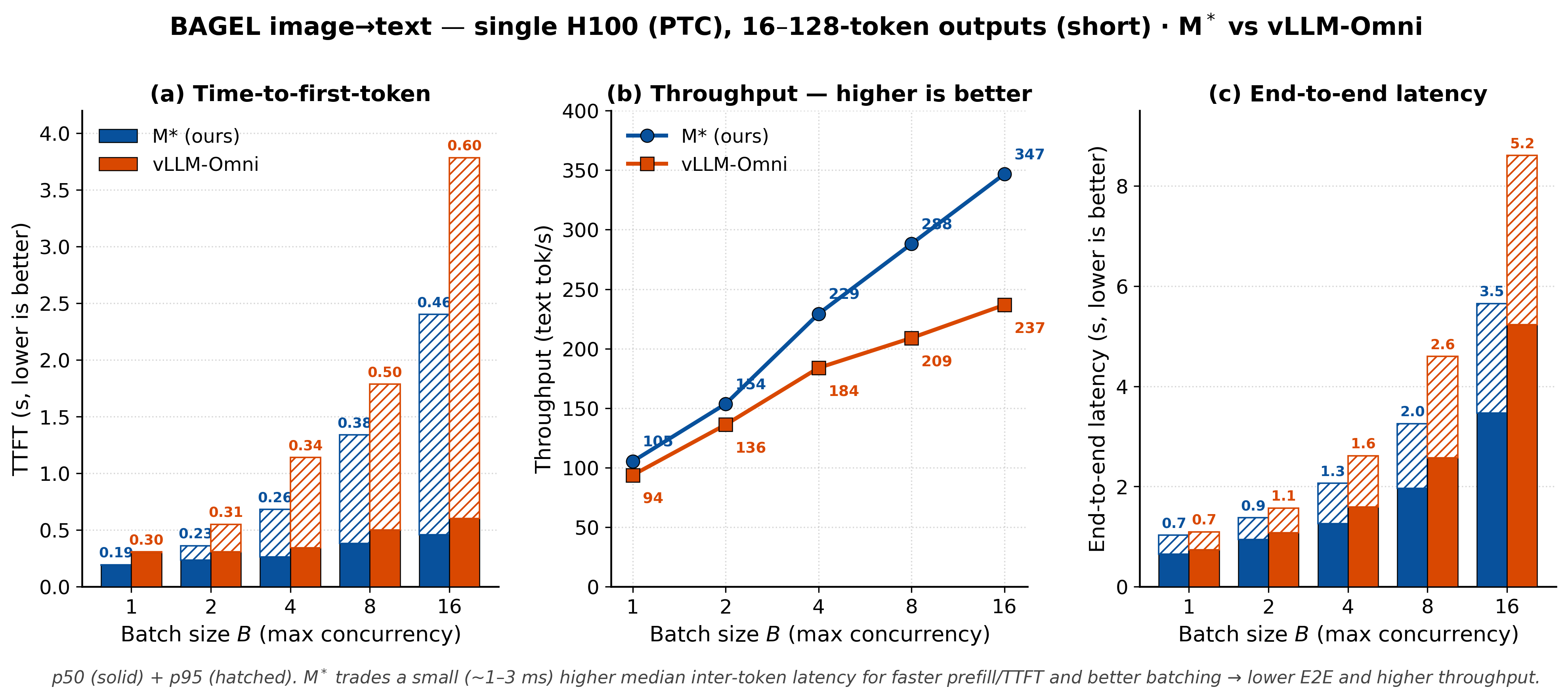}
    \caption{\small\textbf{BAGEL I2T, single H100, $B \in \{1, 2, 4, \dots, 16\}$.}, \texttt{ignore\_eos} with output lengths between 16 and 128. (a) TTFT (log $y$). (b) Throughput (req/s). (c) End-to-end request latency.
    }
    \label{fig:bagel-i2t-1gpu-short}
\end{figure}

\begin{figure}[!htbp]
    \centering
    \includegraphics[width=0.99\linewidth]{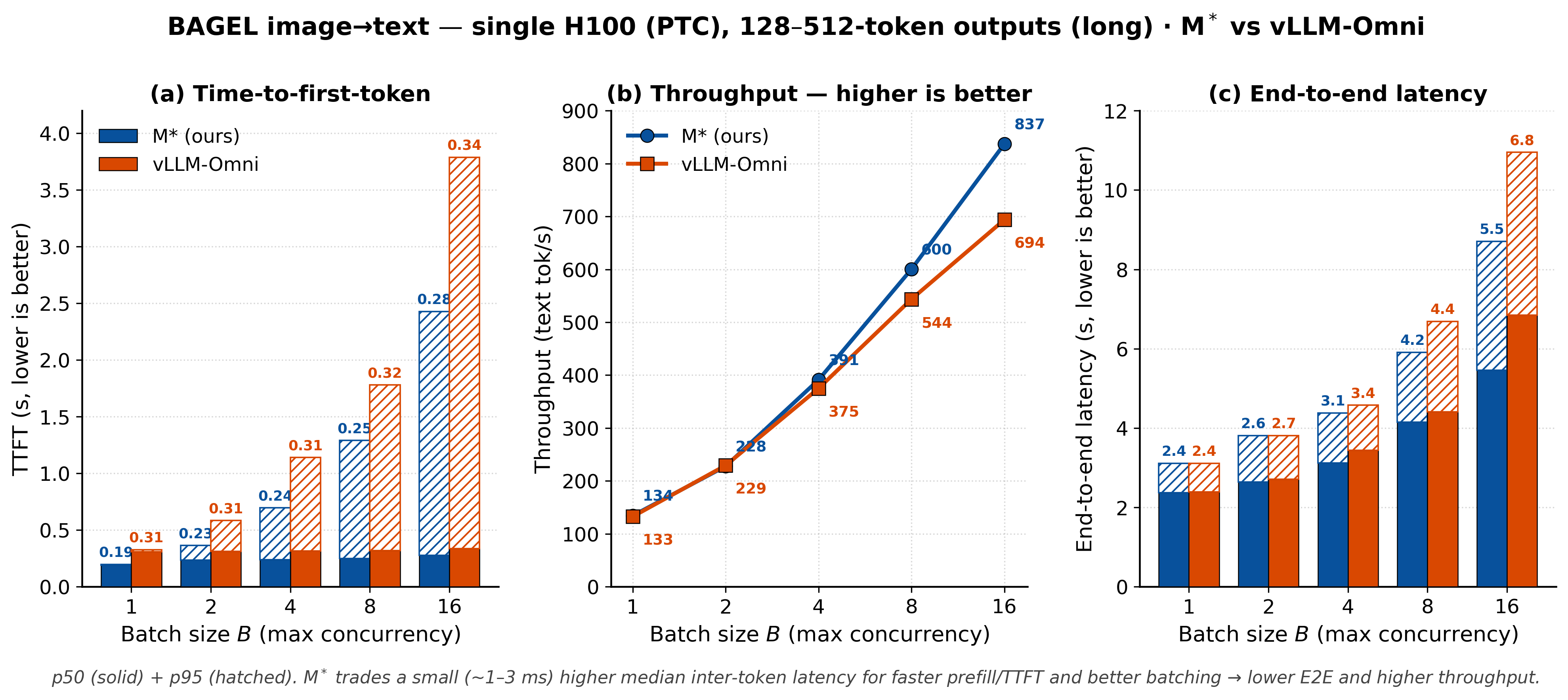}
    \caption{\small\textbf{BAGEL I2T, single H100, $B \in \{1, 2, 4, \dots, 16\}$.}, \texttt{ignore\_eos} with output lengths between 128 and 512. (a) TTFT (log $y$). (b) Throughput (req/s). (c) End-to-end request latency.
    }
    \label{fig:bagel-i2t-1gpu-long}
\end{figure}

Figures~\ref{fig:bagel-i2t-1gpu-short} and \ref{fig:bagel-i2t-1gpu-long} show \sys{} vs. vLLM-Omni on BAGEL I2T for varying output length workloads: one where output lengths are randomly sampled between 16 and 64 tokens, and one where output lengths range from 128 to 512 tokens.
Overall, we have the largest advantage on shorter-decode and higher-concurrency workloads, while remaining comparable in the longer-sequence, lower-concurrency setting.

\FloatBarrier

\section{Walk Graph Primitives: Image Generation Walk with CFG Parallelism}
\label{sec:cfg-example}

\begin{minipage}{\textwidth}
\begin{lstlisting}[style=walkgraph]
image_gen_cfg = Sequential([
  Loop(
    section = Sequential([
      Parallel([                          # 3-branch CFG, 1 branch per rank
        Node("LLM",          [latents, t], [Edge("combine_cfg","v_main")]),
        Node("LLM_cfg_text", [latents, t], [Edge("combine_cfg","v_text")]),
        Node("LLM_cfg_img",  [latents, t], [Edge("combine_cfg","v_img")]),
      ]),
      Node("combine_cfg",
           [v_main, v_text, v_img, latents, t],
           [loopback to all three LLMs])
    ]),
    max_iters = num_timesteps - 1,        # 49 Euler steps
    outputs = [Edge("vae_decoder", "latents")]),
  Node("vae_decoder", [latents], [Edge(EMIT_TO_CLIENT, "image_output")]),
])
\end{lstlisting}
\end{minipage}

\section{Broader Impacts}
\label{app:impacts}
Our work has several potential positive impacts. By improving the efficiency, scalability, and accessibility of multimodal model deployment, the framework can enable broader access to advanced AI capabilities in applications such as accessibility technologies, scientific computing, education, healthcare, and real-time decision support systems. Improved serving efficiency may also reduce infrastructure cost and energy consumption per inference, lowering barriers for smaller organizations and researchers to deploy multimodal AI systems. At the same time, these capabilities may introduce negative societal impacts if misused. More efficient deployment of multimodal models could enable large-scale surveillance, automated misinformation generation, deepfakes, or invasive analysis of visual and audio data. Incorrect model outputs in high-stakes applications may also propagate harmful or biased decisions, particularly if deployed without appropriate human oversight. To mitigate these risks, we emphasize that model authors should deploy such instances with appropriate safeguards, including privacy protections, usage restrictions for sensitive applications, monitoring and auditing mechanisms, and careful human evaluation before deployment in safety-critical settings.

\section{Limitations}
\label{app:limitations}
First, although we evaluate our framework across a diverse set of multimodal models and workloads, it is infeasible to exhaustively test all existing and emerging multimodal architectures, modalities, and deployment settings. As a result, performance characteristics and system behavior may differ for models, serving patterns, or hardware configurations not included in our evaluation. Second, as stated in \Cref{sec:eval}, many evaluation settings that we ran are not included due to significant correctness and performance issues found in baselines, which may have been due to evaluating these systems in previously untested settings. Thus, we did not include these results for fairness. 

In addition, we plan to explore several aspects of the system design in future work. For example, the set of supported runtime engines could be expanded or modified as new models and multimodal tasks emerge, and our current streaming edge policy and edge abstractions represent only a limited number of strategies in a broader design space. Similarly, although \sys{} supports tensor-parallel sharding of graph components, our implementation does not yet explore additional forms of parallelism (e.g., sequence, context, or pipeline parallelism), more advanced asynchronous worker scheduling and modality or model-specific optimization strategies such as sparse attention techniques, or other potential system-level optimizations beyond those described in the paper. These limitations suggest opportunities for future work and further evaluation across a wider range of models, hardware platforms, and distributed serving configurations.

\section{Artifact and Code Reproducibility Details}
\label{app:artifact}

This appendix consolidates the information needed to reproduce the
results in \S\ref{sec:eval}: hardware, software stack, per-experiment
configurations, and the licenses of all third-party code, model
weights, and datasets used.  Configuration files, dockerfiles, and the
exact reproduction commands will accompany the public source release at
the camera-ready deadline.
We commit to working with reviewers to ensure that all results are reproduced before the camera-ready deadline.

\paragraph{Hardware.}
Experiments run on two single-node clusters with intra-node NVLink and
shared-memory (SHM) tensor transport between worker processes; no
inter-node communication is exercised.
\begin{itemize}[leftmargin=1.6em,topsep=0pt,itemsep=0pt]
  \item One $4{\times}$NVIDIA H100 (80~GB) node, used for BAGEL on its
  3-GPU CFG-parallel deployment (one rank per CFG branch, encoders /
  VAE colocated with rank~0) and V-JEPA~2 on a single GPU.
  \item One $8{\times}$NVIDIA H200 (141~GB) node, used for Qwen3-Omni
  on a 2-GPU disaggregated deployment (Thinker on rank~1,
  Talker$+$Code2Wav on rank~0) and Orpheus on a single GPU.
\end{itemize}

\paragraph{Software stack.}
\sys{} requires Python~3.12 and PyTorch with CUDA.  The engine layer
uses FlashInfer~\citep{ye2025flashinfer} for paged-attention prefill
and decode, HuggingFace
Transformers~\citep{wolf2020transformers} for tokenization and weight
loading, and \texttt{torchaudio} / \texttt{torchcodec} for audio and
video decoding.  

\paragraph{Workload configurations.}
Every experiment is driven by our harness: \texttt{num\_warmup} warm-up requests at
sequential concurrency, followed by \texttt{num\_requests} timed
requests in the workload's configured concurrency mode (default:
\texttt{offline}, sized waves of $B$ requests).  Defaults are
\texttt{num\_warmup}=3 and the per-model settings below; greedy
decoding (temperature~=~0) is forced on every sub-model so cross-system
token sequences match.
Each configuration uses 3--5 warmup requests followed by 10--160 timed requests (with the total number of requests being least 5$\times$ the maximum concurrency); we report mean, p50, and p95 wherever appropriate.

\begin{itemize}[leftmargin=1.6em,topsep=0pt,itemsep=0pt]
  \item \textbf{BAGEL-7B-MoT} (\texttt{configs/bagel\_cfg\_parallel.yaml}):
  $1024^2$ output, 50-step rectified-flow schedule,
  \texttt{cfg\_img\_scale}=2 / \texttt{cfg\_renorm\_type=text\_channel}
  on I2I; workloads T2I, I2I, I2T at $B\in\{1,4,8\}$ on VBench prompts.
  \item \textbf{Qwen3-Omni-30B-A3B-Instruct} (\texttt{configs/qwen3omni\_2gpu.yaml}):
  Seed-TTS evaluation set, \texttt{max\_tokens}=256,
  \texttt{thinker/talker/cp\_temperature}=0, system prompt matching
  vLLM-Omni's official Qwen3-Omni examples; $B\in\{1,4,8,16,32\}$.
  \item \textbf{Orpheus-3B} (\texttt{configs/orpheus.yaml}): LLM on
  rank~0, SNAC decoder on rank~1; 3-engine walk
  (Embeddings $\to$ LLM $\to$ SNAC), $B\in\{1,2,4,8,16\}$.
  \item \textbf{V-JEPA~2 ViT-g (AC)} (\texttt{configs/vjepa2\_ac.yaml}):
  DROID episodes, 8 frames per request at $256\times256$, bf16,
  sequential ($B{=}1$), rollout horizons $H\in\{4,15,30\}$.  The
  upstream Meta vjepa2 baseline reproduces the verbatim driver pattern
  from \texttt{notebooks/energy\_landscape\_example.ipynb} (Cells 5--6).
\end{itemize}

\paragraph{Software.}
Table~\ref{tab:license-code} lists the third-party systems used as
baselines or as direct dependencies of \sys{}'s engine layer.  Where a
license could not be confirmed from the official repository, the cell
is left blank.

\begin{table}[h]
  \centering
  \small
  \caption{Software / code licenses.}
  \label{tab:license-code}
  \setlength{\tabcolsep}{6pt}
  \renewcommand{\arraystretch}{1.15}
  \begin{tabular}{@{}l l l@{}}
  \toprule
  \textbf{Asset and version} & \textbf{URL} & \textbf{License} \\
  \midrule
  vLLM-Omni (vllm: v0.21.0) \citep{yin2026vllmomni}      & \url{https://github.com/vllm-project/vllm-omni}          & Apache 2.0 \\
  SGLang-Omni (commit: 4a3960) \citep{sglangomni}                 & \url{https://github.com/sgl-project/sglang-omni}         & Apache 2.0 \\
  VoxServe v0.1.0 \citep{kamahori2026voxserve}   & \url{https://github.com/vox-serve/vox-serve}             & Apache 2.0 \\
  Meta vjepa2 (upstream baseline) \citep{assran2025vjepa2} & \url{https://github.com/facebookresearch/vjepa2}         & Apache 2.0 \\
  BAGEL reference implementation \citep{deng2025bagel} & \url{https://github.com/bytedance-seed/Bagel}            & Apache 2.0 \\
  Orpheus-TTS reference \citep{orpheus2025} & \url{https://github.com/canopyai/Orpheus-TTS}            & Apache 2.0 \\
  openpi ($\pi_{0.5}$ reference) \citep{black2025pi05}                & \url{https://github.com/Physical-Intelligence/openpi}    & Apache 2.0 \\
  SNAC (audio codec) \citep{siuzdak2024snac} & \url{https://github.com/hubertsiuzdak/snac}              & MIT \\
  HuggingFace Transformers \citep{wolf2020transformers}        & \url{https://github.com/huggingface/transformers} & Apache 2.0 \\
  HuggingFace Diffusers \citep{von-platen-etal-2022-diffusers} & \url{https://github.com/huggingface/diffusers}    & Apache 2.0 \\
  FlashInfer \citep{ye2025flashinfer}            & \url{https://github.com/flashinfer-ai/flashinfer}        & Apache 2.0 \\
  Mooncake transfer engine \citep{qin2025mooncake} & \url{https://github.com/kvcache-ai/Mooncake}           & Apache 2.0 \\
  \bottomrule
  \end{tabular}
\end{table}

\paragraph{Model weights.}
Table~\ref{tab:license-models} lists pretrained checkpoints used in the
evaluation, with HuggingFace mirrors (or upstream URLs) and licenses.
Checkpoints are downloaded from these locations on first use; \sys{}
performs no further fine-tuning.  The V-JEPA~2 AC checkpoint is the
encoder$+$action-conditioned predictor bundle published on FAIR S3
(HuggingFace does not host an AC variant).

\begin{table}[h]
  \centering
  \small
  \caption{Model checkpoint licenses.}
  \label{tab:license-models}
  \setlength{\tabcolsep}{6pt}
  \renewcommand{\arraystretch}{1.15}
  \begin{tabular}{@{}l l l@{}}
  \toprule
  \textbf{Asset} & \textbf{URL} & \textbf{License} \\
  \midrule
  BAGEL \citep{deng2025bagel}                  & \url{https://huggingface.co/ByteDance-Seed/BAGEL-7B-MoT}              & Apache 2.0 \\
  Qwen3-Omni \citep{xu2025qwen3omni} & \url{https://huggingface.co/Qwen/Qwen3-Omni-30B-A3B-Instruct}         & Apache 2.0 \\
  Orpheus \citep{orpheus2025}               & \url{https://huggingface.co/canopylabs/orpheus-3b-0.1-ft}             & Apache 2.0 \\
  V-JEPA~2 \citep{assran2025vjepa2}        & \url{https://huggingface.co/facebook/vjepa2-vitg-fpc64-256}         & Apache 2.0 \\
  \bottomrule
  \end{tabular}
\end{table}

\paragraph{Datasets and benchmarks.}
Table~\ref{tab:license-datasets} lists the evaluation datasets.  Where
a license could not be confirmed from the official source, the cell is
left blank.

\begin{table}[h]
  \centering
  \small
  \caption{Dataset / benchmark licenses.}
  \label{tab:license-datasets}
  \setlength{\tabcolsep}{6pt}
  \renewcommand{\arraystretch}{1.15}
  \begin{tabular}{@{}l l l@{}}
  \toprule
  \textbf{Asset} & \textbf{URL} & \textbf{License} \\
  \midrule
  VBench \citep{huang2024vbench} & \url{https://github.com/Vchitect/VBench}                & Apache 2.0 \\
  Seed-TTS \citep{anastassiou2024seedtts} & \url{https://github.com/BytedanceSpeech/seed-tts-eval} & CC BY 4.0 \\
  DROID \citep{khazatsky2024droid} & \url{https://huggingface.co/datasets/lerobot/droid_100} & MIT \\
  \bottomrule
  \end{tabular}
\end{table}

\end{document}